\documentclass[journal]{IEEEtran}

\usepackage{times}
\usepackage{epsfig}
\usepackage{graphicx}
\usepackage{amsmath}
\usepackage{booktabs}
\usepackage{xcolor}
\usepackage{multirow}
\usepackage{amssymb}
\newcommand{\heading}[1]{\noindent\textbf{#1}}

\usepackage[hidelinks]{hyperref}
\hypersetup{
    colorlinks=true,%
    urlcolor=magenta
}

\begin{document}

\title{Geo-Localization via Ground-to-Satellite Cross-View Image Retrieval}

\author{Zelong~Zeng,
        Zheng~Wang,~\IEEEmembership{Member,~IEEE,}
        Fan~Yang,
        and~Shin'ichi~Satoh,~\IEEEmembership{Member,~IEEE}
\thanks{This work was supported by National Key R\&D Program of China (No.2021YFC3320301) and by National Natural Science Foundation of China (No.62171325). \textit{(Corresponding author: Zheng Wang)}}
\thanks{Z.~Wang is with the School of Computer Science, National Engineering Research Center for Multimedia Software, Wuhan University, China. (e-mail: wangzwhu@whu.edu.cn)}
\thanks{Z.~Zeng, F.~Yang, and S.~Satoh are with the Department of Information and Communication Engineering, Graduate School of Information Science and Technology, The University of Tokyo, Japan, and also with the Digital Content and Media Sciences Research Division, National Institute of Informatics, Japan. (e-mail: zzlbz@nii.ac.jp;yang@nii.ac.jp;satoh@nii.ac.jp)}
\thanks{Manuscript received XXX; revised XXX.}}

\markboth{IEEE Transactions on Multimedia}%
{Zeng \MakeLowercase{\textit{et al.}}: Geo-Localization via Ground-to-Satellite Cross-View Image Retrieval}

\maketitle

\begin{abstract}
The large variation of viewpoint and irrelevant content around the target always hinder accurate image retrieval and its subsequent tasks. In this paper, we investigate an extremely challenging task: given a ground-view image of a landmark, we aim to achieve cross-view geo-localization by searching out its corresponding satellite-view images. Specifically, the challenge comes from the gap between ground-view and satellite-view, which includes not only large viewpoint changes (some parts of the landmark may be invisible from front view to top view) but also highly irrelevant background (the target landmark tend to be hidden in other surrounding buildings), making it difficult to learn a common representation or a suitable mapping.

To address this issue, we take advantage of drone-view information as a bridge between ground-view and satellite-view domains. We propose a Peer Learning and Cross Diffusion (PLCD) framework. PLCD consists of three parts: 1) a peer learning across ground-view and drone-view to find visible parts to benefit ground-drone cross-view representation learning; 2) a patch-based network for satellite-drone cross-view representation learning; 3) a cross diffusion between ground-drone space and satellite-drone space. Extensive experiments conducted on the University-Earth and University-Google datasets show that our method outperforms state-of-the-arts significantly.
\end{abstract}

\IEEEpeerreviewmaketitle

\section{Introduction}

\IEEEPARstart{C}{ross-view} {image retrieval consist in retrieving} the most relevant images from different platforms, which has received significant attention in recent years due to a large number of potential applications~\cite{zhang2016sketch,he2016cross,gu2018multi,lu2019efficient,zheng2020adaptive,verma2021unsupervised}. {\cite{zhang2016sketch} proposed a sketch-based image retrieval method, \cite{he2016cross} focused on image-text cross-modal retrieval, \cite{gu2018multi} analysed cross-modal fashion retrieval tasks, \cite{lu2019efficient} proposed a hashing approach for large-scale multimedia search tasks, and \cite{zheng2020adaptive} used a multi-view hashing approach for social image retrieval tasks.} Most previous works attempt to find a feature representation that is robust to the view variations. They have leveraged the deep neural networks together with metric learning strategies to learn discriminative representations for images~\cite{workman2015wide,arandjelovic2016netvlad,hu2018cvm,zheng2020university,vo2016localizing,cai2019ground,radenovic2018fine,shi2020optimal,bai2017ensemble,bai2018regularized}. {They designed different network architectures (\textit{e.g.}, two-streams architectures~\cite{hu2018cvm,zheng2020university,vo2016localizing,cai2019ground}) and aggregated layers (\textit{e.g.}, NetVLAD~\cite{arandjelovic2016netvlad} and GeM~\cite{radenovic2018fine}) for geo-localization task. \cite{bai2017ensemble,bai2018regularized} used manifold diffusion for re-ranking and enhanced the evaluation of image retrieval.} Most of them also involved the spatial alignment~\cite{hu2020multi,NEURIPS2019_ba2f0015,liu2019lending} and the attention mechanism~\cite{cai2019ground} in designing networks. 

\begin{figure}[t]
	\centering
		\includegraphics[width=0.95\linewidth]{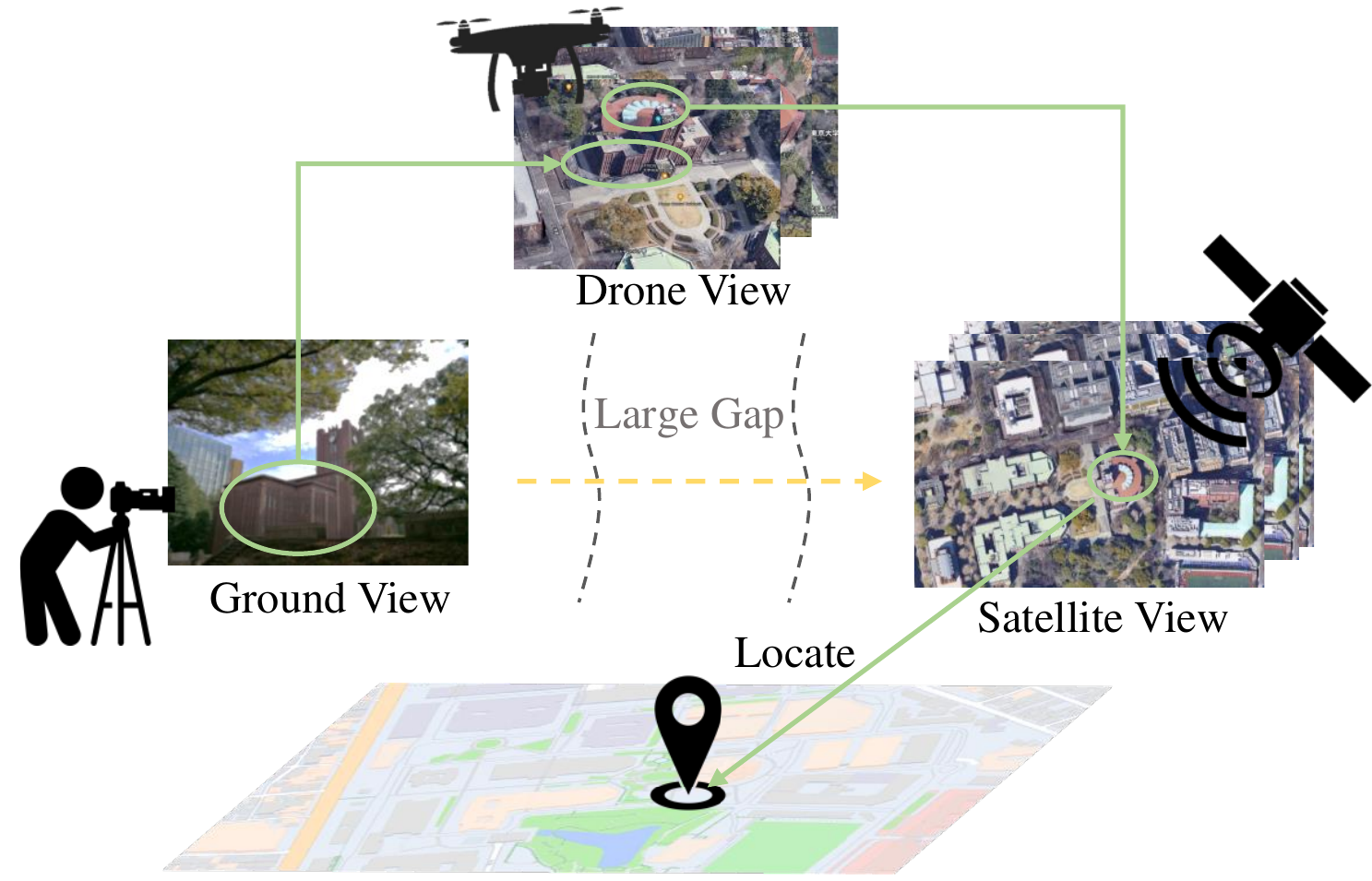} 
   \caption{An illustration of the task: given a ground-view image of a landmark, we aim to achieve cross-view geo-localization by searching out its corresponding satellite-view images. There is a very big gap between ground-view and satellite-view images due to the invisibility of parts of landmark and highly irrelevant background. We introduce the drone-view images to fill this gap and achieve the ground-drone-satellite image retrieval.} 
\label{fig:example}
\end{figure}

In practice, ground-to-satellite image retrieval is beneficial to image-based geo-localization. For instance, given a ground-view image showing the target landmark, the system retrieves and locates the same landmark among candidate satellite-view images. It can facilitate the positioning devices, like GPS, to provide a more accurate localization result. Previous works of ground-to-satellite image retrieval tried to learn a common representation across different views. Besides, two datasets CVUSA~\cite{workman2015localize} and CVACT~\cite{liu2019lending} are widely used for evaluation while the ground-view images provided by them are both panorama images.
We argue that 1) the viewpoint varies greatly between ground-view and satellite-view images, where most components of the landmark shown in the ground-view image are invisible to the satellite-view; 2) it is not practical for users to obtain panorama ground-view images, thus the afore-mentioned datasets are not suitable for the real-world setting.

To tackle these issues, we propose to utilize drone-view images to bridge the domain gap between the ground-view and the satellite-view. As Figure~\ref{fig:example} shows, a drone-view image may share some common components of the landmark with a ground-view image as well as a satellite-view image. This makes it reasonable to learn representations for ground-to-drone and drone-to-satellite respectively. Inspired by this observation, we design a Peer Learning and Cross Diffusion (PLCD) framework, which is an effective way to build a connection from ground-view domain to satellite-view domain. Specifically, we divide the process of ground-to-satellite image retrieval into three steps: 1) ground-to-drone (ground$\rightarrow$drone) image retrieval, 2) drone-to-satellite (drone$\rightarrow$satellite) image retrieval, 3) diffusion between ground-drone and 
satellite-drone spaces. In the first step, considering the label noise (for a ground image, the relative drone-view images do not necessarily share the same facet regions of the landmark, because drone-view images of the same landmark are captured from different drone-viewing directions) and the background noise (the target landmark tend to be hidden in other surrounding buildings), we design a peer learning network which allows ground-view domain and drone-view domain to assist each other, and find the region of the target to benefit learn ground-drone cross-view representation. In the second step, we design a patch-based network to learn satellite-drone representation. For the third step, we generate a cross-view relation graph by using a cross diffusion to connect ground-drone and satellite-drone spaces.  In Section~\ref{section:method}, we introduce each part in more detail. 

To verify the effectiveness of the proposed method, we conduct experiments on two datasets University-Earth and University-Google~\cite{zheng2020university}, PLCD achieves 40.87\%/21.77\% CMC@1\% accuracies for the task of ground-to-drone retrieval which are higher than the baseline work~\cite{zheng2020university} by 34.24\%/16.57\% on the University-Earth and University-Google datasets respectively. Our contributions can be summarized as follows:

\begin{itemize}
\item {\bf A new strategy.} We raise a new strategy for ground-to-satellite image retrieval. The strategy is taking advantage of drone-view information as a bridge between ground-view and satellite-view domains to overcome the huge domain gap.
\item {\bf A novel method.} We propose a novel Peer Learning and Cross Diffusion (PLCD) framework, which builds mapping from ground-view domain to satellite-view domain. The method achieves great performance improvements on the {University-Earth and University-Google datasets}.

\end{itemize}

\section{Related Work}

\subsection{Ground-to-Satellite Geo-Localization}
Ground-to-satellite geo-localization (Ground-view image $\rightarrow$ Satellite-view image) has been attracting more attention in recent years due to a large number of potential applications. It's a task of image localization on a geo-referenced satellite map given a query ground-view image. Workman \textit{et al.}~\cite{workman2015wide} attempted to fine-tune a pre-trained CNN by reducing the feature distance between pairs of ground-view images and satellite-view images for the cross-view localization task. Zhai \textit{et al.}~\cite{zhai2017predicting} designed a modified Siamese Network by plugging the NetVLAD~\cite{arandjelovic2016netvlad} layer, they demonstrated that siamese-structure and aggregation layer can make image descriptors robust against large viewpoint changes. To take one step further, Hu \textit{et al.}~\cite{hu2018cvm} proposed a weighted soft margin ranking loss function that not only speeds up the training convergence but also improved the final matching accuracy. Shi \textit{et al.}~\cite{shi2020optimal} proposed CVFT layer which can transport features from one domain to the other, leading to more meaningful feature similarity comparison. Furthermore, Shi \textit{et al.}~\cite{NEURIPS2019_ba2f0015} improved the performance of ground-to-satellite geo-localization through spatial alignment and attention mechanism. For these existing works, the ground-view images they used are all panorama images, which means each panorama image contains the same area as the relative satellite-view image. Most of the existing methods~\cite{huang2021occluded} focus on how to directly establish the association between images from different views, \textit{i.e.}, to directly map images from different views to the same space. Different from them, our proposed method uses an intermediate view (drone view) to reduce the impact of large gaps between different views. In addition, we propose to use cross-diffusion to better connect the embedding spaces of different views.

\begin{figure*}[t]
\centering
\includegraphics[width=\linewidth]{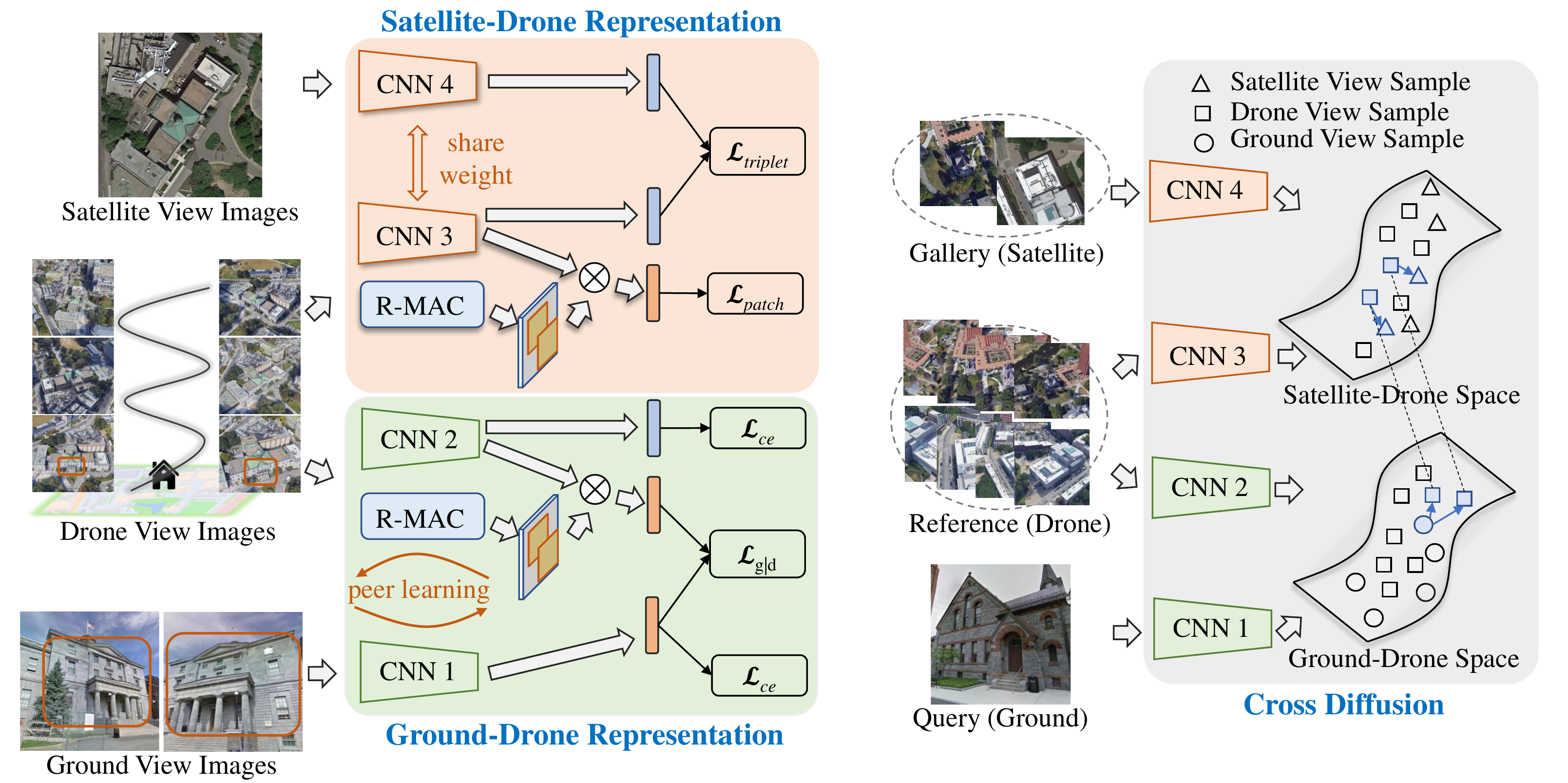}
\caption{The framework overview of our method. It consists of three parts, \textit{i.e.}, 1) peer learning for the ground-drone cross-view representation; 2) patch-based model for the satellite-drone cross-view representation learning; 3) cross diffusion between ground-drone and satellite-drone spaces. We first learn the networks $\textrm{CNN}_1$ and $\textrm{CNN}_2$ for the ground-drone representation, as well as $\textrm{CNN}_3$ and $\textrm{CNN}_4$ for the satellite-drone representation. Then given a ground-view image of a landmark, cross diffusion is capable of achieving cross-view geo-localization by searching out corresponding satellite-view images, where we also use drone-view images (\textsl{no label}) acting as the reference set.}
\label{fig:framework}
\end{figure*}

\subsection{Drone-to-Satellite Geo-Localization}
Drone-to-satellite geo-localization (Drone-view image $\rightarrow$ Satellite-view image) is a new task of cross-view geo-localization. Given one drone-view image or video, the task aims to find the most similar satellite-view image to localize the target landmark in the satellite view. Because of the demand for drone applications, {for instance, multi-view coordination~\cite{olszewska2016interest}, drone-view coordination~\cite{olszewska2017clock},} Drone-to-satellite geo-localization has been attracting more attention.

Zheng \textit{et al.}~\cite{zheng2020university} proposed Drone-to-satellite geo-localization task and a new multi-view multi-source benchmark for drone-based geo-localization, named University-1652. They demonstrated that the University-1652 helps the model to learn the viewpoint-invariant features and also has good generalization ability in the real-world scenario. Hu \textit{et al.}~\cite{hu2020multi} proposed an orientation-based method to align the patterns and introduce a new branch to extract aligned partial feature\cite{sun2018beyond}. Moreover, they provided a style alignment strategy to reduce the variance in image style and enhance the feature unification. Inspired by the human visual system, Wang \textit{et al.}~\cite{wang2021each} proposed a model which explicitly takes contextual patterns into consideration and leverages the surrounding environment around the target landmark, yielding better performance.

Some methods align images with different views at the image level, but their proposed methods require human pre-processing of the image, such as rotating the image to the same direction~\cite{hu2020multi} or cropping the image~\cite{wang2021each}. In addition, most methods are based on global-level feature. Considering the presence of noisy samples, our approach uses peer learning, which allows the model to be adaptively align during training to extract the best matching local features and obtain better results.

\subsection{Cross-Modal Image Retrieval}
Cross-modal image retrieval allows using other types of query~\cite{zhong2021grayscale,yang2020instance,kansal2020sdl}. Some of the common types are text-to-image retrieval and sketch-to-image retrieval.
\begin{itemize}
    \item \textbf{Text-to-image retrieval.} Most cross-model IR on text-to-image retrieval considers the case where the input query is formulated as an input image plus text describing the required modifications to the image. The main challenge of this task is to fuse the modified textual information into the source image and to evaluate the similarity between the fused image and the features of the desired image. Most approaches have focused on designing feature fusion layers and learning loss functions using different deep metrics. TIRG~\cite{vo2019composing} used ResNet and RNN (LSTM) to extract features of the source image and modified text respectively, then fused them by using a gated residual connection, and finally used metric learning to learn the similarity metric. \cite{guo2018dialog} extracted the features of the source image and that of the text by ResNet and RNN (GRU), and fused them through concatenation, then applied reinforcement learning to improve the quality of similarity metric learning.
    \item \textbf{Sketch-to-image retrieval.} The sketch-based image retrieval problem involves retrieving a hand-drawn sketch of a given object from a gallery of images of that class of objects. To bridge the gap between the sketch domain and the image domain, most approaches take the form of extracting the edge map of the image first, and then feeding the edge map and sketch into a deep network structure to extract features and perform matching. \cite{pang2017cross} used generative networks to enhance the discrimination of the extracted features. \cite{song2017deep} applied the attention mechanism module and proposes a new loss function, namely HOLEF, to improve the performance.
\end{itemize}

In contrast to text/sketch-to-image retrieval, the data in the Ground-to-Satellite cross-view Image Retrieval (G2S IR) task are all images and no text/sketch data, so information fusion and information alignment for different modalities do not need to be considered in our work. Our work focuses more on feature mapping of multi-view images.

\section{Our Method}
\label{section:method}
In this section, we introduce the Peer Learning and Cross Diffusion (PLCD) framework (Figure~\ref{fig:framework}). Due to the gap between ground-view and satellite-view, too large to cover by a common space, PLCD prepares ``doable'' common spaces for ground-drone and drone-satellite independently, and connect them via cross diffusion. PLCD framework can be decomposed into three main parts: 1) A peer learning across ground-view and drone-view to find visible correlative parts to enjoy benefit ground-drone local representation learning (Section~\ref{subsection:peering learning}). {The peer learning includes two steps, \textit{i.e.,} train the senior peer and train the junior peer.} 2) A patch-based model to obtain global representation across drone-view and satellite-view, and close the gap between ground-drone and drone-satellite spaces (Section~\ref{subsection:patch-based}). 3) A cross diffusion to best utilize intermediate view, \textit{i.e.}, drone view, for boosting ground-to-satellite retrieval as post processing (Section~\ref{subsection:cross diffusion}).

\heading{Problem formulation.} For ground-drone-satellite image retrieval, we have three subsets containing images of three different views, \textit{i.e.}, ground-view, drone-view and satellite-view. We use $\mathcal{G}=\left\{\mathbf{g}\right\}$, $\mathcal{D}=\left\{\mathbf{d}\right\}$, and $\mathcal{S}=\left\{\mathbf{s}\right\}$ to denote the ground-view, drone-view, and satellite-view sets, respectively. For each image $\mathbf{g}$, $\mathbf{d}$, or $\mathbf{s}$, we use $y \in [1,\ldots,C]$ to represent its label, where $C$ is the number of identities. Note that there are only identity labels and no annotated bounding box indicating the region of the landmark. The sets $\mathcal{G}$, $\mathcal{D}$ and $\mathcal{S}$ are divided into two parts, one part for training, the other for testing, and the training and testing parts have no overlapping identity. Given an image $\mathbf{g}$, our aim is to search out corresponding images that contain the same landmark in the set $\mathcal{S}$. In this paper, we learn cross-view representations via four CNN models. They are 1) the ground-drone representation model $\textrm{CNN}_1$ for ground-view images $f_{\theta_1}(\mathbf{g}): \mathcal{G} \mapsto \mathbb{R}^{d}$; 2) the ground-drone representation model $\textrm{CNN}_2$ for drone-view images $f_{\theta_2}(\mathbf{d}): \mathcal{D} \mapsto \mathbb{R}^{d}$; 3) the satellite-drone representation model $\textrm{CNN}_3$ for drone-view images $f_{\theta_3}(\mathbf{d}): \mathcal{D} \mapsto \mathbb{R}^{d}$; and 4) the satellite-drone representation model $\textrm{CNN}_4$ for satellite-view images $f_{\theta_4}(\mathbf{s}): \mathcal{G} \mapsto \mathbb{R}^{d}$, where $d$ stands for the number of dimensions. 

\subsection{Ground-Drone Cross-view Representation}
\label{subsection:peering learning}
In major public benchmarks, each image is generally associated with a landmark identity, here called hard label. However, for each ground image, corresponding drone-view images do not necessarily share the same facet regions of the landmark. This is because drone-view images of the landmark are captured from different directions. If we train the model by directly using hard labels, the performance will not be so good. However, hard labels can help to discover potential valuable images that benefit training.

\begin{figure}[t]
	\centering
		\includegraphics[width=0.85\linewidth]{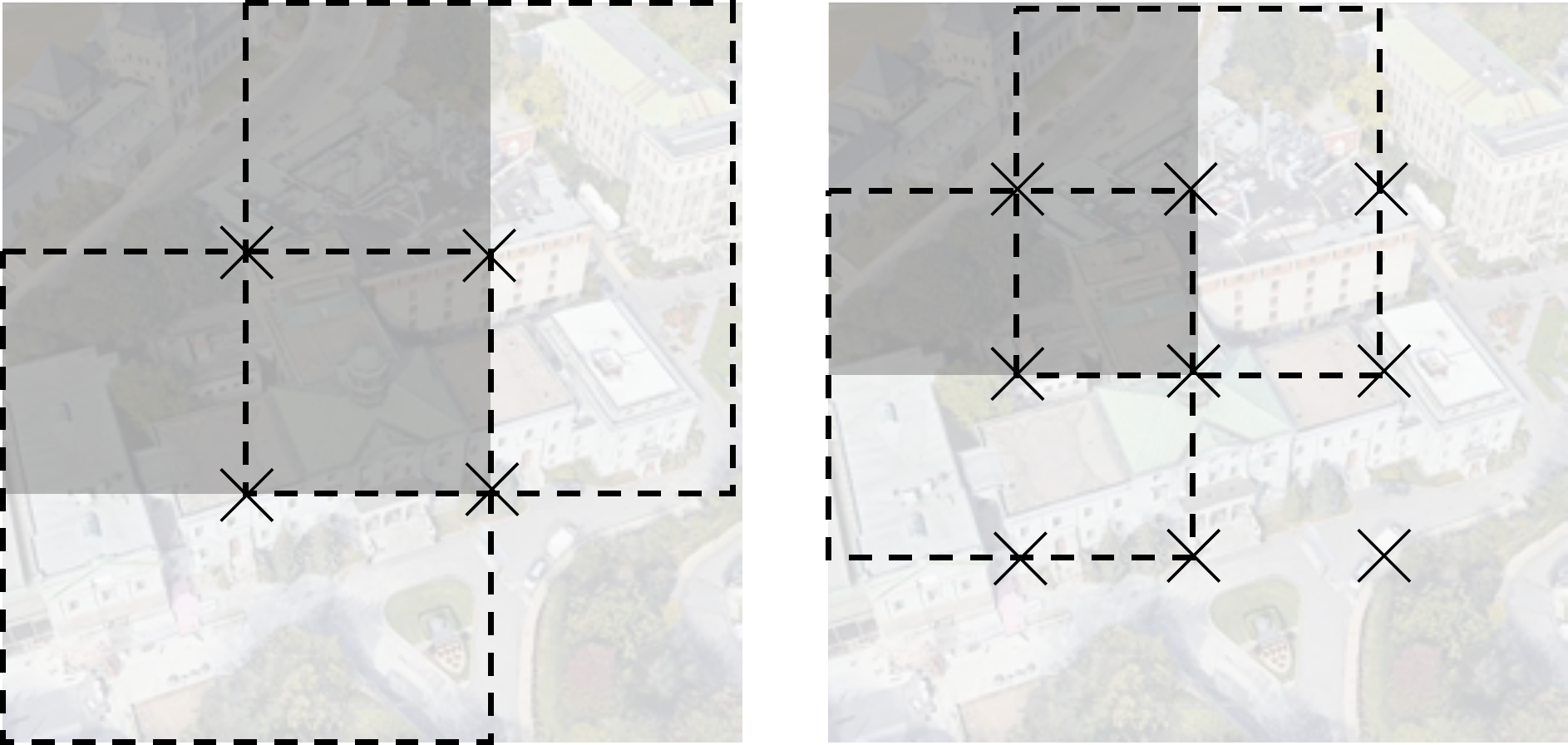} 
   \caption{An illustration of R-MAC layer. The figure shows example regions extracted at two different scales ($l$ = 2, 3), the top-left region of each scale (gray colored region) and its neighboring regions (dashed borders). The centers of all regions are with a cross. At scale $l$, we sample $l \times l$ regions with the width $L/(l + 1)$, where $L$ indicates the length of square image.} 
\label{fig:r-mac}
\end{figure}

To make full use of valuable positive samples and focus on precise regions of targets, we propose a peering learning network which contains two branches, \textit{i.e.}, the ground-view branch and the drone-view branch. Two branches do not share weights, but their final embedding features are pushed close to each other by constraints. Given a ground-view image, the \textit{Detection} layer (RPN~\cite{ren2015faster} pretrained on~\cite{gordo2016deep}) outputs the region of interest and the $\textrm{CNN}_1$ outputs the feature map. Next, a \textit{Region of Interest} (ROI) pooling layer is applied to the detection result and outputs the feature. For the drone-view branch, the $\textrm{CNN}_2$ is used to extract activation features from drone-view images. Activation features are max-pooled in different regions of the image by using a multi-scale rigid grid with overlapping cells, called \textit{R-MAC}~\cite{noh2017large} (as illustrated in Figure~\ref{fig:r-mac}) layer. We use cross-entropy to optimize each branch respectively, and a consistency loss to learn the relation between two domains. The training process can be divided into two steps. 

\subsubsection{Step I: Train the Senior Peer}
\label{subsection:step1}
To avoid the effect from hard label noise, we first use the easiest samples. 
We feed `easy' triplets as explained below to $\textrm{CNN}_1$ and $\textrm{CNN}_2$. Each triplet consists of one ground-view image $\mathbf{g}$, its easiest positive image $\mathbf{d}^{*}$ (share the same facet region of the landmark as $\mathbf{g}$) and its multiple difficult negative images $\{\mathbf{d}_{j}|^N_{j=1}\}$. Such triplets are sampled according to the image features $f_{\theta_1}$ or $f_{\theta_2}$. The easiest positive image $\mathbf{d}^*$ is the top-1 ranking drone-view image from a positive image batch. To ensure that there exist easy positive images in each batch, we divide the drone-view images into $D$ different sections according to drone's directions and randomly select one positive image from each section to form a batch. The most difficult negative images $\{\mathbf{d}_{j}|^N_{j=1}\}$ are the top-N ranking gallery image from a negative image batch. In this process, two branches assist each other and learn how to select the easiest positive and difficult negative samples. Here, we adopt a consistency loss~\cite{liu2019stochastic} to project features from different domains to the same space. It is formulated as:

\begin{small} 
\begin{equation}
\label{eq:softmax-based-loss}
\mathcal{L}_{\mathbf{g} \mid \mathbf{d}}\!(\theta_1, \!\theta_2\!)\!\!= \!\!
\frac{\exp \!\left(\!-\left\|f_{\theta_1}(\mathbf{g})\!-\!f_{\theta_2}(\mathbf{d}^*)\right\|^{2}\!\right)\!}
{\exp \!\!\left(\!\!-\!\!\left\|f_{\theta_1}(\mathbf{g})\!\!-\!\!f_{\theta_2}(\mathbf{d}^*)\right\|^{2}\!\!\right)
\!\!
+
\!\!\sum_{j=1}^{N} \!\!\exp \!\!\left(\!\!-\!\!\left\|f_{\theta_1}(\mathbf{g})\!\!-\!\!f_{\theta_2}(\mathbf{d}_j)\right\|^{2}\!\!\right)\!},
\end{equation}
\end{small}
where $\left \|\ \cdot\ \right \|$ denotes the Euclidean Norm.

We also apply cross-entropy loss to improve the discriminative representation. The objective loss of step I $\mathcal{L}_{hard}$ combines the $\mathcal{L}_{\mathbf{g} \mid \mathbf{d}}$ and cross-entropy loss as:

\begin{equation}
\label{eq:hard-loss}
\mathcal{L}_{hard} = \mathcal{L}_{\mathbf{g} \mid \mathbf{d}} - \sum_{j=1}^{C}p_{j}\operatorname{log}q_{j},
\end{equation}
where $q_j$ is a score obtained by applying a softmax function to the logits of a classification layer for the sample, $p_j$ is the ground truth for the sample, and $C$ denotes the number of classes.

\subsubsection{Step II: Train the Junior Peer}
\label{subsection:step2}

In step I, we only use the easiest positive samples and global information. To make full use of difficult positive samples and pay more attention to local information, we use the trained models of step I as a senior peer to guide a junior peer in step II. The structure of the junior peer is the same as the senior peer. Two peers both activate \textit{R-MAC} layers.

We denote $\theta_{1}^{SP}$ and $\theta_{2}^{SP}$ are parameters of $\textrm{CNN}_1$ and $\textrm{CNN}_2$ in senior peer models, and we freeze them in step II. $\theta_1^{JP}$ and $\theta_2^{JP}$ are learnable parameters of $\textrm{CNN}_1$ and $\textrm{CNN}_2$ in junior peer models. Both of them are fed with doublets, and each doublet consists of one ground-view image $\mathbf{g}$, its multiple positive images $\{\mathbf{d}_{i}|^M_{i=1}\}$. Given a positive drone-view image $\mathbf{d}_i$, $\textrm{CNN}_2$ and \textit{R-MAC} layer can output descriptors $\{f_{\theta_2}(\mathbf{d}_{i,1}), \cdots, f_{\theta_2}(\mathbf{d}_{i,m})\}$ corresponding to the $m$ potential image sub-regions $\{\mathbf{d}_{i,1}, \cdots, \mathbf{d}_{i,m}\}$ proposed by \textit{R-MAC} layer. The similarities between $\mathbf{g}$ and sub-regions are measured by dot products and normalized by a softmax function with temperature $\tau$ over the $M$ positive images in the batch:

\begin{equation}
\begin{split}
&\mathcal{S}\left(\mathbf{g}, \mathbf{d}_{1}, \cdots, \mathbf{d}_{M}; \tau \right) = \operatorname{softmax} ([\left\langle f_{\theta_{1}}(\mathbf{g}), f_{\theta_2}(\mathbf{d}_{1})\right\rangle / \tau,  \\
&\left\langle f_{\theta_{1}}(\mathbf{g}),  f_{\theta_2}(\mathbf{d}_{1,1})\right\rangle / \tau, \cdots, \left\langle f_{\theta_{1}}(\mathbf{g}), f_{\theta_2}(\mathbf{d}_{1,m})\right\rangle / \tau, \cdots,\\
&\left\langle f_{\theta_{1}}(\mathbf{g}), f_{\theta_2}(\mathbf{d}_{M})\right\rangle / \tau,\left\langle f_{\theta_{1}}(\mathbf{g}), f_{\theta_2}(\mathbf{d}_{M,1})\right\rangle / \tau, \cdots,\\
&\left\langle f_{\theta_{1}}(\mathbf{g}), f_{\theta_2}(\mathbf{d}_{M,m})\right\rangle / \tau]),
\end{split}
\label{eq:softmax-func}
\end{equation}
where $f_{\theta_1}(\mathbf{g})$ is the encoded feature representations of the ground-view image by the $\textrm{CNN}_1$ branch, $f_{\theta_2}(\mathbf{d}_{i})$ and $f_{\theta_2}(\mathbf{d}_{i,j})$ are the encoded feature representations of the $i$-th gallery image and its $j$-th potential sub-region by the $\textrm{CNN}_2$ branch respectively. $\tau$ is temperature hyper-parameter less than $1$, which makes the similarity vector sharper.

$\mathcal{S}^{SP}$ and $\mathcal{S}^{JP}$ are the similarities calculated by Equation~\ref{eq:softmax-func} of senior peer model and junior peer model respectively. $\mathcal{S}^{SP}$ shows the distribution of the drone-view image and its sub-regions in senior peer model space, which serves as soft supervision to junior peer model via a cross-entropy loss:

\begin{equation}
\label{eq:soft-loss}
\mathcal{L}_{soft} = 
\!-\!\!\sum_i\! \!\mathcal{S}^{SP}_i\!\!\left(\!\mathbf{g}, \mathbf{d}_{1},\!\cdots\!,\mathbf{d}_{M}; \tau \!\right)\! \log \!\left ( \mathcal{S}^{JP}_i\!\!\left(\!\mathbf{g}, \mathbf{d}_{1},\!\cdots,\!\mathbf{d}_{M}; 1 \!\right)\! \right),
\end{equation}
where $\mathcal{S}_i(\cdot)$ is the $i$-th element of softmax vector $\mathcal{S}(\cdot)$. At the same time, we also use hard loss $\mathcal{L}_{hard}$ (Equation~\ref{eq:hard-loss}) to supervise the junior peer model. $\mathcal{L}_{hard}$ and $\mathcal{L}_{soft}$ are jointly adopted in step II as:

\begin{equation}
\label{eq:joint-loss}
\mathcal{L_{G-D}} = \mathcal{L}_{hard} + \lambda_1 \mathcal{L}_{soft},
\end{equation}
where the weights $\lambda_1$ is used for balancing these two losses.

\subsection{Satellite-Drone Cross-view Representation}
\label{subsection:patch-based}
We design a patch-based network, which has two weight shared branches: $\textrm{CNN}_3$ and $\textrm{CNN}_4$. $\textrm{CNN}_3$ and $\textrm{CNN}_4$ can extract feature maps $\mathbf{m}_d$ and $\mathbf{m}_s$ from input drone-view image and satellite-view image respectively. Here, we use the pretrained peer learning model to supervise patch-based network for training, we found that this process can improve the performance of diffusion. Similar with we describe in Section~\ref{subsection:step2}, given a positive drone-view image $\mathbf{d}_i$, $\textrm{CNN}_2$ (from the pretrained peer learning model and we freeze it) and $\textrm{CNN}_3$ output the feature map $\mathbf{m}_{d_i}^{\theta_2}$ and $\mathbf{m}_{d_i}^{\theta_3}$ respectively. Then $\textit{R-MAC}$ layer proposes sub-regions from each feature map and transform each patch into features $\{f_{\theta_2}(\mathbf{d}_{i,1}), \cdots, f_{\theta_2}(\mathbf{d}_{i,m})\}$ and $\{f_{\theta_3}(\mathbf{d}_{i,1}), \cdots, f_{\theta_3}(\mathbf{d}_{i,m})\}$ respectively. This process can be written as:

\begin{equation}
\label{eq:patch-feature-extraction}
\{f_{s}(\mathbf{d}_{i,j})|j\in\{1,\cdots,m\}\} = \mathcal{F}_{\text{R-MAC}}(\mathbf{m}_{d_i}^s), s\in\{\theta_2,\theta_3\},
\end{equation}
where $\mathcal{F}_{\text{R-MAC}}$ indicates \textit{R-MAC} layer, $m$ is the number of sub-region proposed by \textit{R-MAC} layer from each feature map. 

Then we close the distance between each pair $\{f_{\theta_2}(\mathbf{d}_{i,j}),f_{\theta_3}(\mathbf{d}_{i,j})\}$ by Mean Squared Error (MSE) loss $\mathcal{L}_{mse}$:

\begin{equation}
\label{eq:mseloss}
\mathcal{L}_{patch} = \frac{1}{m}\sum_{i}\sum_{j=1}^{m}\mathcal{L}_{mse}(f_{\theta_2}(\mathbf{d}_{i,j}),f_{\theta_3}(\mathbf{d}_{i,j})),
\end{equation}
where $i$ indicates $i$-th drone-view image of the image batch. 

Besides, we adopt a semi-hard triplet loss~\cite{schroff2015facenet} $\mathcal{L}_{triplet}$ to project features from different domains to the same space. The objective function for this phase can be formulated as:

\begin{equation}
\label{eq:joint-loss2}
\mathcal{L_{S-D}} = \mathcal{L}_{triplet} + \lambda_2 \mathcal{L}_{patch},
\end{equation}
where the weights $\lambda_2$ is used for balancing these two losses.

\begin{figure*}[!h]
    \centering
        \includegraphics[width=0.8\linewidth]{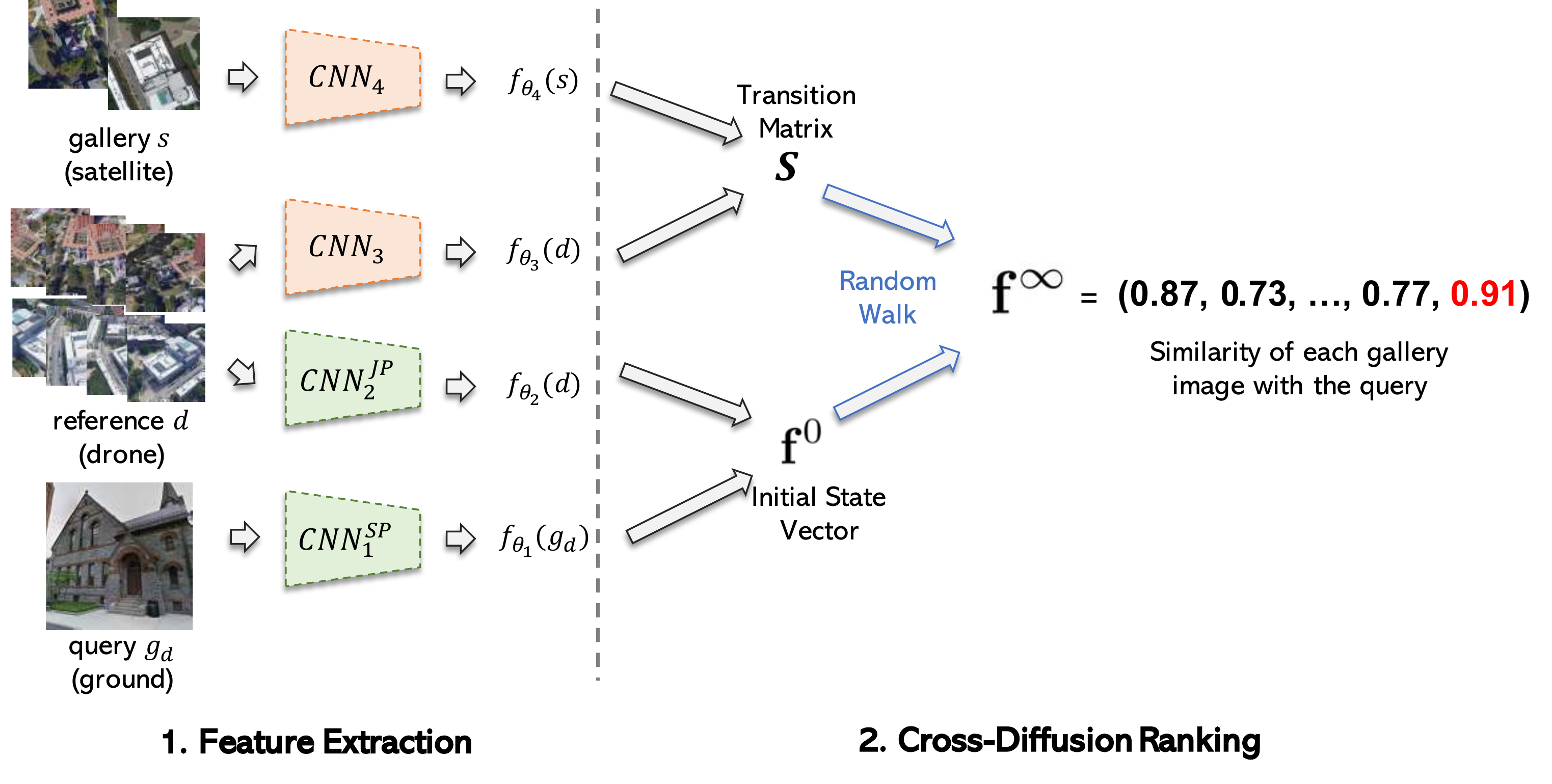}
    \caption{Diagram to show the flow of querying process. It consists of two steps: 1) feature extraction and 2) cross-diffusion ranking.  The output $\mathbf{f^{\infty}}$ corresponds to the satellite-view images as clues for ranking. The higher the weight means that the corresponding satellite-view gallery image is more similar to the drone-view query image.}
    \label{fig:querying}
\end{figure*}

\subsection{Cross-diffusion}
\label{subsection:cross diffusion}
 
Once we obtained the feature representations, we need to connect the ``doable'' common spaces. Some prior works~\cite{bai2018regularized, bai2017ensemble, yang2020mining} have proven the effectiveness of diffusion when mining on manifolds. Here, we apply diffusion, called cross-diffusion, on heterogeneous manifolds (corss-domain) to retrieve the target in satellite-view images with a given ground-view image across the ground-drone and satellite-drone feature spaces.

As a graph-based random walk processing, the key of diffusion is the graph construction and the initialization.
Specifically, the graph construction focuses on the local structure of feature space and defines the direction of random walk, while the initialization determines where to start the random walk.
For the graph, we measure the similarities among image features $\{f_{\theta_3}(\mathbf{d})\}\cup\{f_{\theta_4}(\mathbf{s})\}$ in the satellite-drone feature space, which are used to form a transition matrix $\mathbf{S}\in\mathbb{R}^{(\#\mathcal{D}+\#\mathcal{S})\times(\#\mathcal{D}+\#\mathcal{S})}$ representing the graph.
Essentially, the $i\textrm{-th}$ column of the transition matrix records the probabilities that how likey the weight on the $i\textrm{-th}$ node should diffuse to each of its neighbors, where the node stands for the image sample in our case.
Now that we have obtained the graph for satellite-drone features, we conduct the initialization step by assigning weights to drone-view nodes who are nearest neighbors of the given ground-view query in the ground-drone feature space.
The weights are stored in a state vector $\mathbf{f}\in\mathbb{R}^{\#\mathcal{D}+\#\mathcal{S}}$ whose initial state is defined as
\begin{equation}
\mathbf{f}^0_i =
	\begin{cases}
		(f_{\theta_1}(\mathbf{g}_q)^\top f_{\theta_2}(\mathbf{d}_i))^\gamma & i\in\textrm{NN}^\textrm{ID}(f_{\theta_1}(\mathbf{g}_q)) \\
		0 & \textrm{otherwise}
	\end{cases},
\end{equation}
where $\mathbf{g}_q$ stands for the ground-view query, $\textrm{NN}^\textrm{ID}(\cdot)$ refers to the indexes of nearest neighbors, and $\gamma$ is a constant variable which is set to 3 by convention.
Given the initial state vector and the trainsition matrix, the random walk iterates the following step:
\begin{equation}
	\mathbf{f}^{t+1} = \alpha \mathbf{S} \mathbf{f}^{t} + (1-\alpha) \mathbf{f}^0,\quad \alpha\in(0,1),
\end{equation}
which mathmatically converges to
\begin{equation}
	\mathbf{f}^\infty \propto (\mathbf{I}-\alpha\mathbf{S})^{-1}\mathbf{f}^0
\end{equation}
After convergence, we take the weights in $\mathbf{f}^\infty$ which correspond to the satellite-view images as clues for ranking.

{
\subsection{Querying Process}
The querying process can be divided in to two steps (as shown in Figure~\ref{fig:querying}): 
}

{
\textbf{Feature Extraction.} We use $\textrm{CNN}_{1}^{JP}$ to extract the features $f_{\theta_{1}}(\mathbf{g_q})$ of street-view query image $\mathbf{g_q}$, $\textrm{CNN}_{2}^{JP}$ and $\textrm{CNN}_{3}$ to extract the features $f_{\theta_{2}}(\mathbf{d_i})$ and $f_{\theta_{3}}(\mathbf{d_i})$ of each drone-view reference image $\mathbf{d_i}$, and then $\textrm{CNN}_{4}$ to extract the features $f_{\theta_{4}}(\mathbf{s_j})$ of each satellite-view gallery image $\mathbf{s_j}$. 
}

{
\textbf{Cross-diffusion Ranking.} We adopt diffusion to connect the ``doable'' common spaces (street-drone and drone-satellite). We first use features $f_{\theta_{3}}(\mathbf{d})$ and $f_{\theta_{4}}(\mathbf{s})$ to form the transition matrix $\mathbf{S}$. Next, we initialize the state vector $\mathbf{f^0}$. Finally, given the initial state vector $\mathbf{f^0}$ and the transition matrix $\mathbf{S}$, we adopt random walk to get the weight $\mathbf{f^{\infty}}$ which correspond to the satellite-view images as clues for ranking. The higher the weight means that the corresponding satellite-view gallery image is more similar to the drone-view query image. 
}

\section{Experiments}
\subsection{Dataset and Evaluation Metric}
\heading{Dataset.} \textbf{University-1652}~\cite{zheng2020university} is a recently released multi-view multi-source dataset containing ground-view, drone-view and satellite-view data. It covers 1652 architectures of 72 universities around the world. The training set includes 701 buildings of 33 universities, and the test set includes the other 951 buildings of the rest 39 universities. There are no overlapping universities in the training and test set. Different from other similar multi-view datasets~\cite{workman2015localize,liu2019lending}, the ground-view images of University-1652 are not panorama ones. For each landmark, there are one satellite image, 54 drone-view images from different viewpoints and altitudes and several ground-view images from two different sets (one collected by google earth engine and another by google image search engine). {Note that \cite{zheng2020university} has discussed the limitation of \textbf{University-1652}:
\begin{itemize}
\item[-] To collect images of landmarks, they first obtained metadata about university buildings from Wikipedia (\textit{e.g.}, building name and university affiliation) and then encoded the building names into accurate geographical locations via Google Maps, \textit{i.e.}, latitude and longitude. When we search for locations using Google Maps~\cite{toman2014algorithm}, Google Maps cannot always output precise results. Hence, they filter out buildings whose search results are unclear.
\item[-] Street-view images are collected from Google Maps (Google Street) and the image search engine (Google Image). When images are collected from the image search engine, the collection contains a large number of noisy results, including indoor environments and duplicate images. They applied a model trained on the Place dataset to detect the noisy results and filter them out. Due to accessibility, some buildings do not have good street-view photos, meaning that most street view images are captured by cameras on top of cars. As a result, there are occluded and irrelevant background street view images. Therefore, we applied the RPN~\cite{ren2015faster} detection layer to extract the target regions in the proposed framework.
\end{itemize}
}

We divided the University-1652 dataset into two separate datasets: the first one, called \textbf{University-Earth}, consists of ground images collected by google earth, drone images collected at low altitude and and satellite images; the second one, called \textbf{University-Google}, consists of ground images collected by google image search engine, drone images collected at high altitude and satellite images. Since ground images and drone images are collected in different ways and from different altitudes, they can be considered as collected from different domains, and we usually use Google satellite image for positioning, so both datasets share the same satellite set. We believe that this setting is realistic.

\tabcolsep=3pt
\begin{table*}[t] 
\caption{The comparison with state-of-the-arts in the setting of Ground $\rightarrow$ Satellite retrieval. The CMC@1, CMC@5, CMC@10, CMC@1\% and mAP values (\%) are reported.}
\centering
\label{tab:sota}
\resizebox{1\linewidth}{!}{
\begin{tabular}{l| c c c c c| c c c c c}
\toprule
\multirow{2}{*}{\textbf{Method}} & \multicolumn{5}{c|}{\textbf{University-Earth}} & \multicolumn{5}{c}{\textbf{University-Google}} \\
&{\textbf{CMC@1}} & {\textbf{CMC@5}}  &{\textbf{CMC@10}} & {\textbf{CMC@1\%}} &{\textbf{mAP}} & {\textbf{CMC@1}} & {\textbf{CMC@5}}  &{\textbf{CMC@10}} & {\textbf{CMC@1\%}} &{\textbf{mAP}}\\
\midrule
\midrule
DELF~\cite{noh2017large} \textit{w/o} $\mathcal{D}$& 0.01& 0.39& 0.66& 0.74& 0.60 & 0.01& 0.19& 0.37& 0.37& 0.31\\
DELF~\cite{noh2017large} & 0.12& 0.50& 0.93& 0.93& 0.87& 0.05& 0.26& 0.53& 0.55& 0.53\\
R-MAC~\cite{tolias2015particular} \textit{w/o} $\mathcal{D}$& 1.09& 3.61& 6.59& 6.94& 2.19 & 0.74 & 2.79 & 5.23 & 5.54 & 1.66\\
R-MAC~\cite{tolias2015particular} & 1.09&  3.84& 6.67& 7.06& 2.22& 0.78 & 3.02 & 5.62 & 6.17 & 1.78\\ 
Str-CNN~\cite{zheng2020university} \textit{w/o} $\mathcal{D}$& 0.74&2.79& 4.85&5.66 &1.70 & 0.27& 1.90& 3.76& 4.07& 1.09\\ 
Str-CNN~\cite{zheng2020university} & 1.01&3.22& 6.01&6.63 &2.08 & 0.54& 2.40& 4.77& 5.20& 1.47\\ 
Str-CNN~\cite{zheng2020university} + Multi-loss & 1.51&5.39& 9.77&10.55 &3.12 &1.09 & 4.11 &6.51 &7.13 & 2.28\\ 
CVM-Net~\cite{hu2018cvm} \textit{w/o} $\mathcal{D}$& 0.35&1.05 &2.09 &2.29 &0.88 & 1.24& 3.61& 6.01& 6.24& 2.26\\ 
CVM-Net~\cite{hu2018cvm} & 1.78&4.69 &8.61 &9.42 &3.18 & 0.70& 4.15& 6.36& 6.79& 1.89\\ 
Siam-FCANet50~\cite{cai2019ground} \textit{w/o} $\mathcal{D}$ & 0.39& 2.02& 3.84& 4.23& 1.34 & 0.27 & 1.67 & 2.91 & 3.14 & 0.97\\ 
Siam-FCANet50~\cite{cai2019ground} &1.20 & 4.07& 7.25& 7.68& 2.46 & 1.01 & 3.53 & 6.05 & 6.63 & 2.15\\ 
LPN~\cite{wang2021each} \textit{w/o} $\mathcal{D}$&0.16 & 0.78& 1.82 & 2.06& 0.65 & 0.19 & 0.93 & 1.67 & 1.78 & 0.69\\ 
LPN~\cite{wang2021each} &0.74 & 3.10& 4.69& 5.04& 1.70 & 0.62 & 2.60 & 4.54 & 5.04 & 1.57\\ 
Instance Loss~\cite{zheng2020dual} \textit{w/o} $\mathcal{D}$& 0.62& - & 5.51 & - & 1.60 & -& - & - & - & -\\
Instance Loss~\cite{zheng2020dual} & 1.20 & - & 7.56 & - & 2.52& -& - & - & - & -\\
\midrule
PLCD (Ours) & \textbf{9.15} & \textbf{27.66} & \textbf{38.83} & \textbf{40.87} & \textbf{14.16}& \textbf{4.26} & \textbf{10.62} & \textbf{20.40} & \textbf{21.77} &  \textbf{7.63}\\
\bottomrule
\end{tabular}}
\end{table*}

\heading{Evaluation Metric.} To indicate the performance, the standard Cumulative Matching Characteristics@K (CMC@K) values is adopted. CMC@K represents whether the correctly matched images in the top-K of the ranking list. A higher CMC@K score shows a better performance of the network. We also use mean Average Precision (mAP), which reflects the precision and recall rate of the retrieval performance. Especially, we focus on CMC@K performance on ground-drone image retrieval task, because of existing hard label noise in drone-view images and we only expect to find out those drone-view images share the same facet as the query image.

\subsection{Implementation Details}

In PLCD, we adopt the ResNet-50~\cite{he2016deep} pre-trained on ImageNet~\cite{deng2009imagenet} as $\textrm{CNN}_1$, $\textrm{CNN}_2$, $\textrm{CNN}_3$ and $\textrm{CNN}_4$. We remove the last pooling layer and original classifier for ImageNet, then add one new classifier module. The new classifier module contains five layers: a max-pooling layer $\rightarrow $ a fully connected layer ($\textrm{FC}_1$) $\rightarrow $ a batch normalization layer ($\textrm{BN}$) $\rightarrow $ a dropout layer ($\textrm{Dropout}$) $\rightarrow $ a fully connected layer ($\textrm{FC}_2$). The max-pooling layer output 2048-dim feature and the $\textrm{FC}_1$ project it to 512-dim. All metric learning in our work are based on 512-dim representations. We resize each input image to a fixed size of 384 $\times$ 384 pixels during training and testing. We set $\tau=0.1$, and $L=\{1,2,3,4\}$ which means $\textrm{R-MAC}$ layer can output the potential sub-regions with multi-scale $12 \times 12$, $9 \times 9$, $7 \times 7$ and $5 \times 5$. The model is trained by stochastic gradient descent with momentum 0.9. The learning rate is 0.01 for the classifier module and 0.001 for the rest layers. Dropout rate is 0.5. And $\lambda_1$ and $\lambda_2$ are all set to 1. We train our model for 120 epochs, and the learning rate is decayed by 0.1 after 40 epochs. During testing, we utilize the cosine similarity to measure the similarity between the query image and candidate images in the gallery.

\subsection{Comparison with State-of-the-arts}
Regarding to the ground $\rightarrow$ satellite retrieval, we compared our method, with several typical state-of-the-arts on the University-Earth and University-Google datasets respectively, including DEep Local Features (DELF)~\cite{noh2017large}, R-MAC~\cite{tolias2015particular}, Strong CNN features (Str-CNN)~\cite{zheng2020university}, CVM-Net~\cite{hu2018cvm}, Siam-FCANet50~\cite{cai2019ground}, LPN~\cite{wang2021each}, Instance Loss~\cite{zheng2020dual}, \textit{etc.} Table.~\ref{tab:sota} shows the results. Note that the notation with a `\textit{w/o} $\mathcal{D}$' suffix means that the indicated model has only been trained with the ground and satellite sets. If `\textit{w/o} $\mathcal{D}$' is not included, it means that the model has been trained with ground, drone, and satellite sets. The results clearly demonstrate that our method PLCD has achieved $40.87\%$/$21.77\%$ CMC@1$\%$ accuracy and $14.16\%$/$7.63\%$ mAP on the University-Earth/University-Google dataset on ground-to-satellite (ground $\rightarrow$ satellite) image retrieval.  
The proposed method achieves the best performance and outperforms the state-of-the-arts with a very large margin, for example, our method is higher than the second best method (Str-CNN + Multi-loss) by $30.32\%$/$14.64\%$ and $11.04\%$/$5.25\%$ on CMC@1$\%$ and mAP respectively.

\tabcolsep=1pt
\begin{figure}[t]
	\centering
		\begin{tabular}{l|cccc}
		\footnotesize{Query\ }&\multicolumn{3}{l}{\footnotesize{Ground$\rightarrow$Drone (Top1 $\rightarrow$ Top4)}} & \\
		\includegraphics[width=0.19\linewidth]{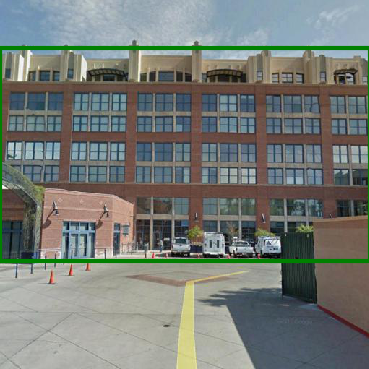} &
		\includegraphics[width=0.19\linewidth]{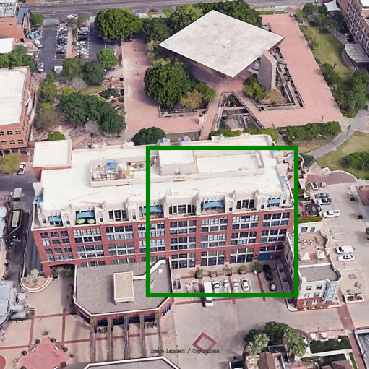} &
		\includegraphics[width=0.19\linewidth]{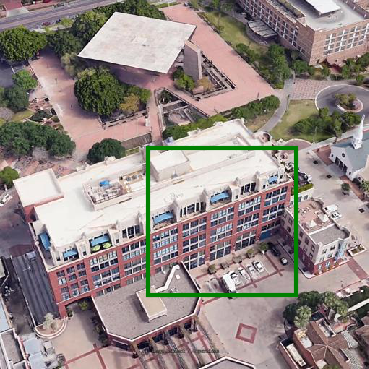} &
		\includegraphics[width=0.19\linewidth]{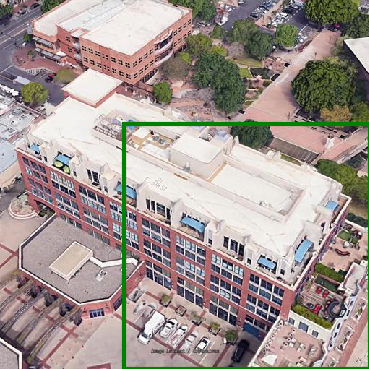} &
		\includegraphics[width=0.19\linewidth]{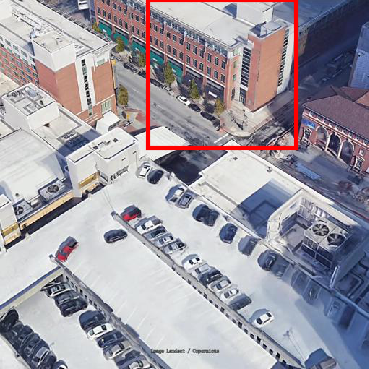} \\
		\includegraphics[width=0.19\linewidth]{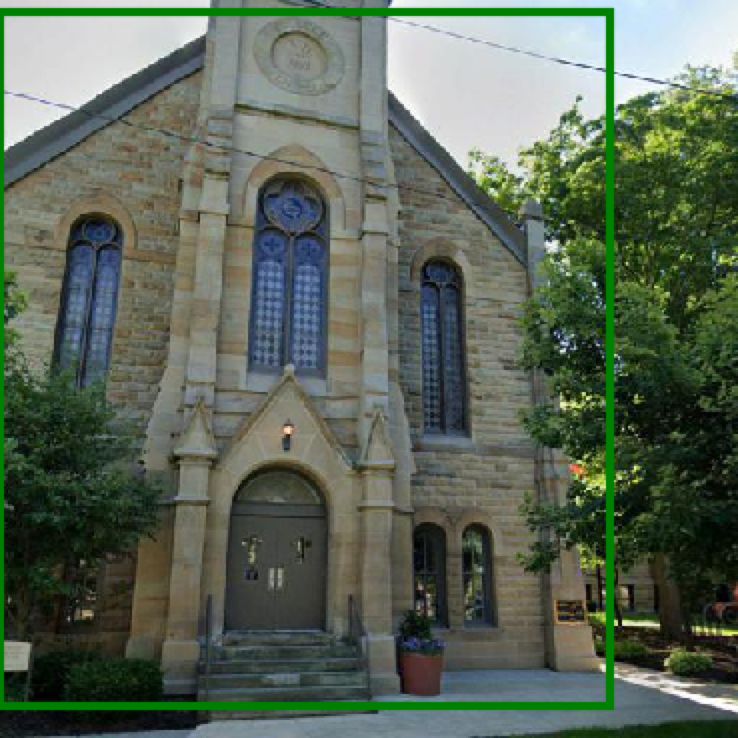} &
		\includegraphics[width=0.19\linewidth]{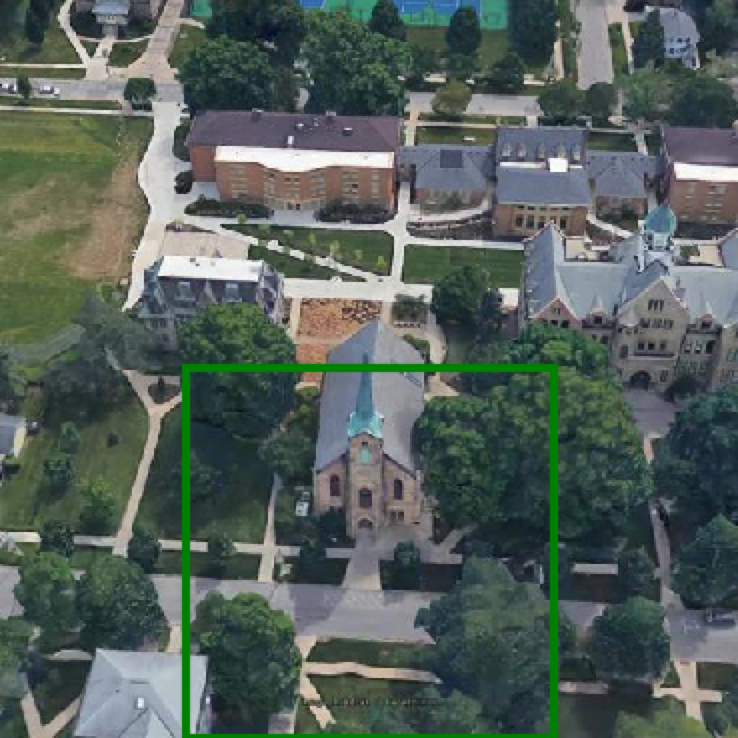} &
		\includegraphics[width=0.19\linewidth]{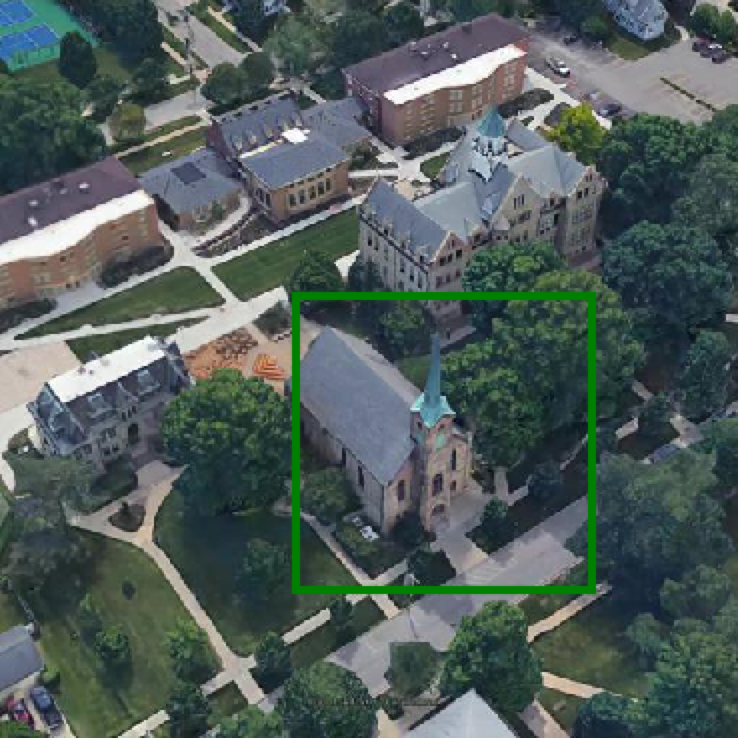} &
		\includegraphics[width=0.19\linewidth]{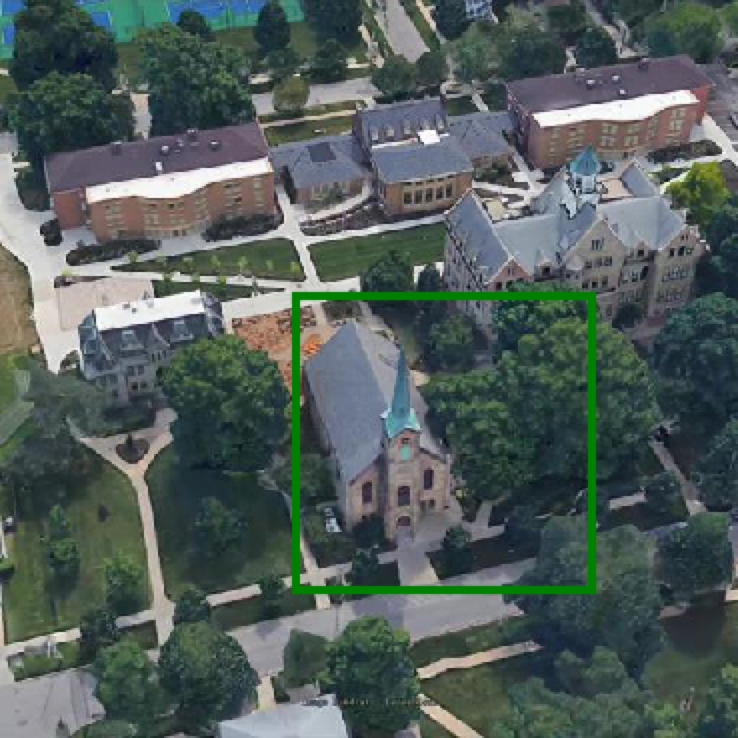} &
		\includegraphics[width=0.19\linewidth]{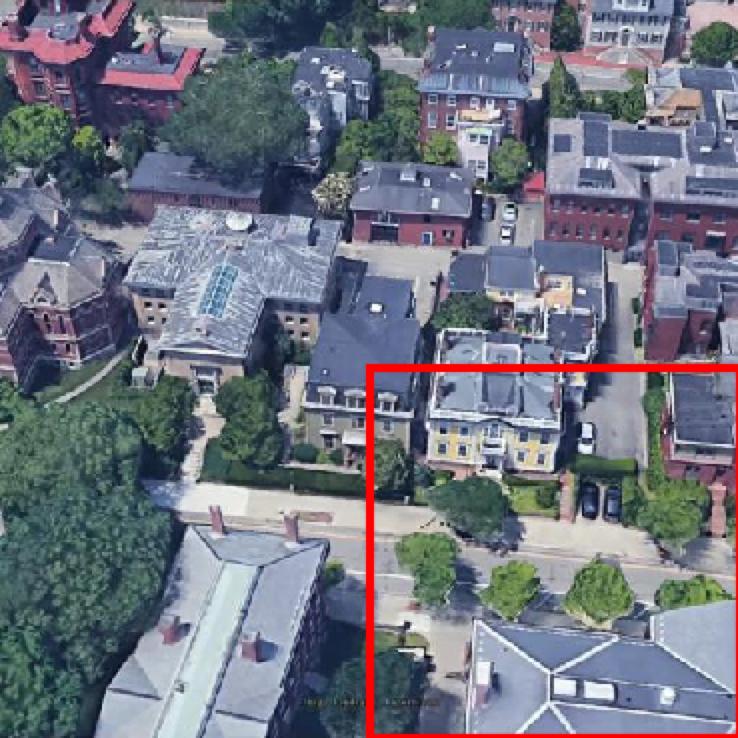} \\
       \end{tabular}
   \caption{The visualization of ground-to-drone image retrieval by peer learning model on the University-Earth dataset. From left to right: ground-view query image and the Top 1-4 retrieved drone-view images. Green and red borders indicate correct and incorrect retrieved results, respectively. Note that peer learning model can find out the target sub-region.} 
\label{fig:visualization}
\end{figure}

\tabcolsep=1.5pt
\begin{table}[t] 
\caption{The comparison with state-of-the-arts in the setting of Ground $\rightarrow$ Drone retrieval. 
}
\centering
\label{tab:sota-drone}
\resizebox{1\linewidth}{!}{
\begin{tabular}{l ccccc}
\toprule
{\textbf{Method}} & {\textbf{CMC@1}} & {\textbf{CMC@5}}  &{\textbf{{CMC@10}}} & \textbf{CMC@1\%} &\textbf{mAP} \\
\midrule
\midrule
DELF~\cite{noh2017large} & 0.04& 0.27& 0.70&14.97 & 1.40\\
R-MAC~\cite{tolias2015particular} &2.09 &3.99  &5.12 &21.79 & 2.34\\ 
Str-CNN~\cite{zheng2020university} &1.51 &3.26 & 4.96 &18.65 &1.65 \\ 
Str-CNN+Semi-hard~\cite{schroff2015facenet} &2.83 &6.05 &8.45 &29.93 & 3.09\\ 
Str-CNN+Semi-hard+GeM~\cite{radenovic2018fine} &2.44  &5.51  & 7.95& 30.21& 3.04\\ 
CVM-Net~\cite{hu2018cvm} & 2.79&5.12 &7.02 &27.18 &2.59 \\ 
Siam-FCANet50~\cite{cai2019ground}  &3.14 & 4.89& 6.48& 24.89& 3.07\\ 
Ours & \textbf{4.77} & \textbf{9.46} & \textbf{13.65} & \textbf{45.99} & \textbf{6.32} \\
\bottomrule
\end{tabular}}
\end{table}

\tabcolsep=2pt
\begin{table}[t] 
\caption{Ablation study of peer learning. The two-branch models are the base network. `S' and `J' stand for the steps of training the senior and junior peer, respectively. `B' represents selecting the best sub-region.}
\centering
\label{tab:ablation-peer}
\resizebox{1\linewidth}{!}{
\begin{tabular}{l ccccc}
\toprule
{\textbf{Method}} & {\textbf{CMC@1}} & {\textbf{CMC@5}}  &{\textbf{{CMC@10}}} & \textbf{CMC@1\%} &\textbf{mAP} \\
\midrule
\midrule
Two-branch &3.14 &6.17 & 8.18 &33.35 &3.48 \\ 
Two-branch+S &3.61 &7.13 &10.00 &37.61 & 3.26\\ 
Two-branch+S+J &4.61  &8.92  & 12.18& 40.44& 3.65\\
Two-branch+S+J+B &\textbf{4.77}  &\textbf{9.46}  & \textbf{13.65} & \textbf{45.99} & \textbf{6.32}\\
\bottomrule
\end{tabular}}
\end{table}

\subsection{The effectiveness of Peer Learning}

To evaluate the effectiveness of peer learning, we conduct some experiments on the University-Earth dataset. we first compared our method with state-of-the-arts on ground-to-drone image retrieval task (ground $\rightarrow$ drone), including DEep Local Features (DELF)~\cite{noh2017large}, R-MAC~\cite{tolias2015particular}, Strong CNN features (Str-CNN)~\cite{zheng2020university}, CVM-Net~\cite{hu2018cvm}, and Siam-FCANet50~\cite{cai2019ground}. Table.~\ref{tab:sota-drone} shows the results. The results demonstrate that our peer learning methods get the best performance compared with all other methods. It shows the effectiveness of the proposed peer learning network.

Second, we conduct an ablation study for the two steps of proposed peer learning. The base network for ground-drone representation exploits a structure of two-branch models. Senior and Junior are the two steps of peer learning strategy. B represents the feature of the best sub-region. It means that, for each gallery image, we use the feature of its best sub-region as its representation. We conduct experiments by adding them one by one. We show the ground-to-drone image retrieval performance in Table.~\ref{tab:ablation-peer}. The results show that each design for peer learning is very important and effective for peer learning.

Moreover, we show some retrieval results on Figure~\ref{fig:visualization}. We observe that our model can retrieve the reasonable image and its sub-region based on the content. Some failure cases are also shown in the fourth column of drone-view image list. We notice that it is challenging in that the irrelevant drone image has a similar pattern with the ground image.

To provide more evidence of the effectiveness of our method, we also visualize the feature distributions. We sample some ground-view images in the testing set of the University-Earth dataset and select their top-50 retrieval results. Figure~\ref{fig:distribution} (a)-(b) show the visualization distribution. Compared with Str-CNN~\cite{zheng2020university}+Multi-loss, the distributions show our peer learning can achieve a better distribution on ground-drone cross-view space and find more positives.

\vspace{-2mm}

\tabcolsep=3pt
\begin{table}[t] 
\caption{Comparison of different mapping methods between ground-drone and satellite-drone domains.}
\vspace{-2mm}
\centering
\label{tab:diffusion}
\resizebox{1\linewidth}{!}{
\begin{tabular}{l ccccc}
\toprule
{\textbf{Method}} & {\textbf{CMC@1}} & {\textbf{CMC@5}}  &{\textbf{{CMC@10}}} & \textbf{CMC@1\%} &\textbf{mAP} \\
\midrule
\midrule
Peer Learning &0.97 &4.73 & 7.99 &8.72 &2.43 \\ 
Projection &0.85 &2,68 & 4.38 &8.30 &1.72 \\ 
Supervision &2.36 &7.08 &11.69 &12.34 & 3.82\\ 
Diffusion &\textbf{9.15}  &\textbf{27.66}  & \textbf{38.38}& \textbf{40.87}& \textbf{14.16}\\

\bottomrule
\end{tabular}}
\end{table}

\tabcolsep=2pt
\begin{table}[t] 
\caption{Comparison of one common model and our two-branch model in different settings.  
}
\vspace{-3mm}
\centering
\label{tab:one common model}
\resizebox{1\linewidth}{!}{
\begin{tabular}{l ccccc}
\toprule
{\textbf{Method}} & {\textbf{CMC@1}} & {\textbf{CMC@5}}  &{\textbf{{CMC@10}}} & \textbf{CMC@1\%} &\textbf{mAP} \\
\midrule
\midrule
Ground$\rightarrow$Drone \\
One Model  &3.33 &6.98 & 9.15 &34.90 &4.49 \\ 
Two-branch &\textbf{4.77}  &\textbf{9.46}  & \textbf{13.65} & \textbf{45.99} & \textbf{6.32}\\
\midrule
Drone$\rightarrow$Satellite  \\
One Model  &63.95 &77.29 & 80.30 &80.71 &66.96\\ 
Two-branch &\textbf{67.06} &\textbf{82.84} & \textbf{87.23} &\textbf{87.69} &\textbf{70.60} \\ 
\midrule
Ground$\rightarrow$Satellite \\
One Model   &6.62 &17.77 & 26.29 &27.76 &10.23 \\ 
Two-branch  & \textbf{9.15} & \textbf{27.66} & \textbf{38.83} & \textbf{40.87} & \textbf{14.16}\\ 
\bottomrule
\end{tabular}}
\end{table}

\vspace{-1mm}

\subsection{The effectiveness of Cross Diffusion}

To show the effectiveness of the proposed diffusion methods. We compared with different mapping methods to connect the ground-drone and satellite-drone domains. The mapping methods include: 1) Peer Learning: applying peer learning strategy to train a common space for ground-drone-satellite representation; 2) Projection: training a new fully connective (FC) layer to project drone-view features of ground-drone and satellite-drone models to a same space; 3) Supervision: when training the drone-satellite model, for each drone-view image, we close the distance between its features extracted by the drone-satellite model and the pretrained ground-drone model; 4) Diffusion: the proposed cross diffusion method. The the ground-satellite image retrieval performances by these four methods are shown in Table.~\ref{tab:diffusion}. The results clearly indicate that our diffusion method is the most effective method to connect the ground-drone and satellite-drone domains. 

\begin{figure*}[t]
    \centering
    \tabcolsep=1pt
    \begin{tabular}{ccccc}
	    \includegraphics[width=0.2\linewidth]{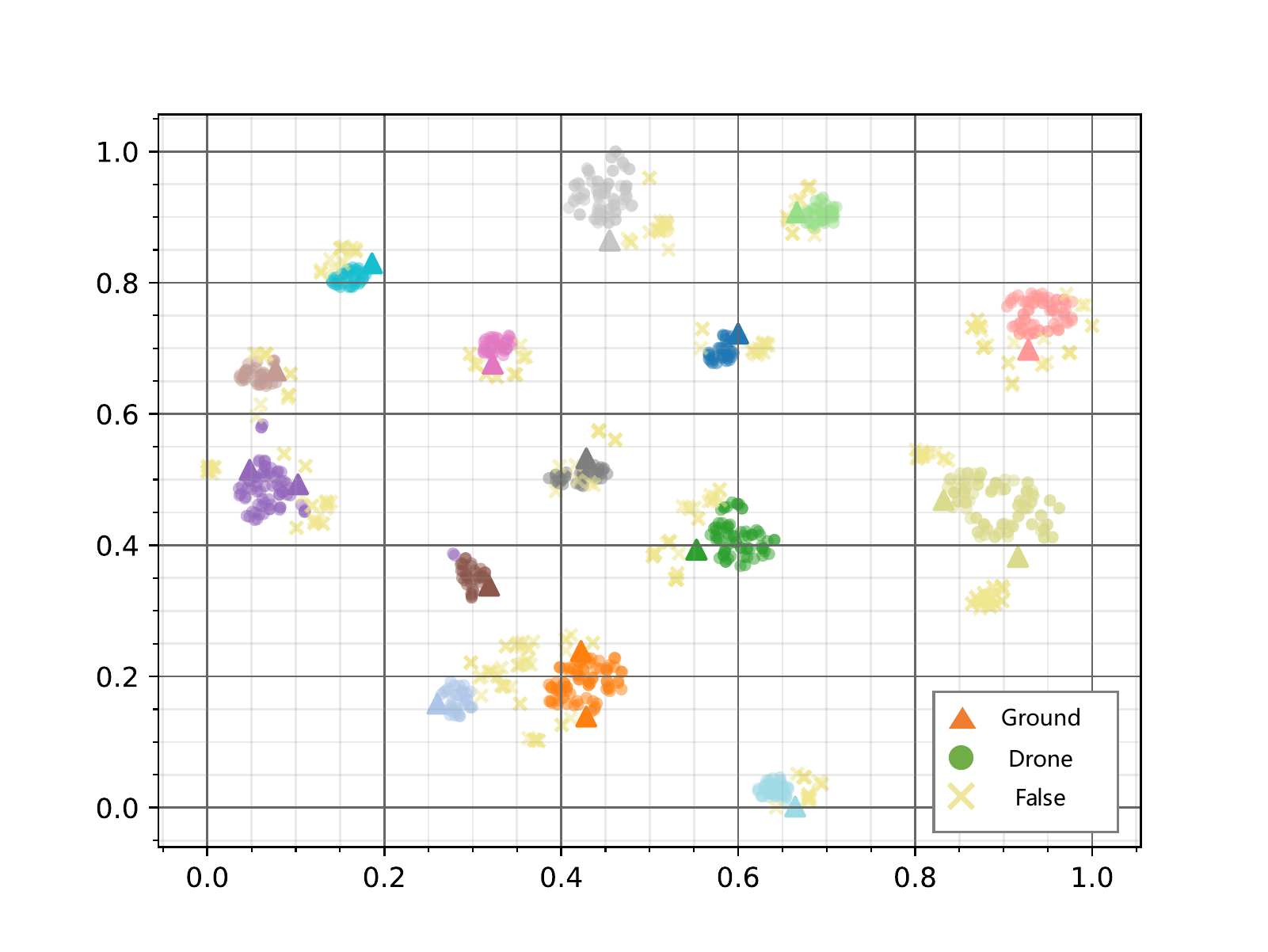} &
	    \includegraphics[width=0.2\linewidth]{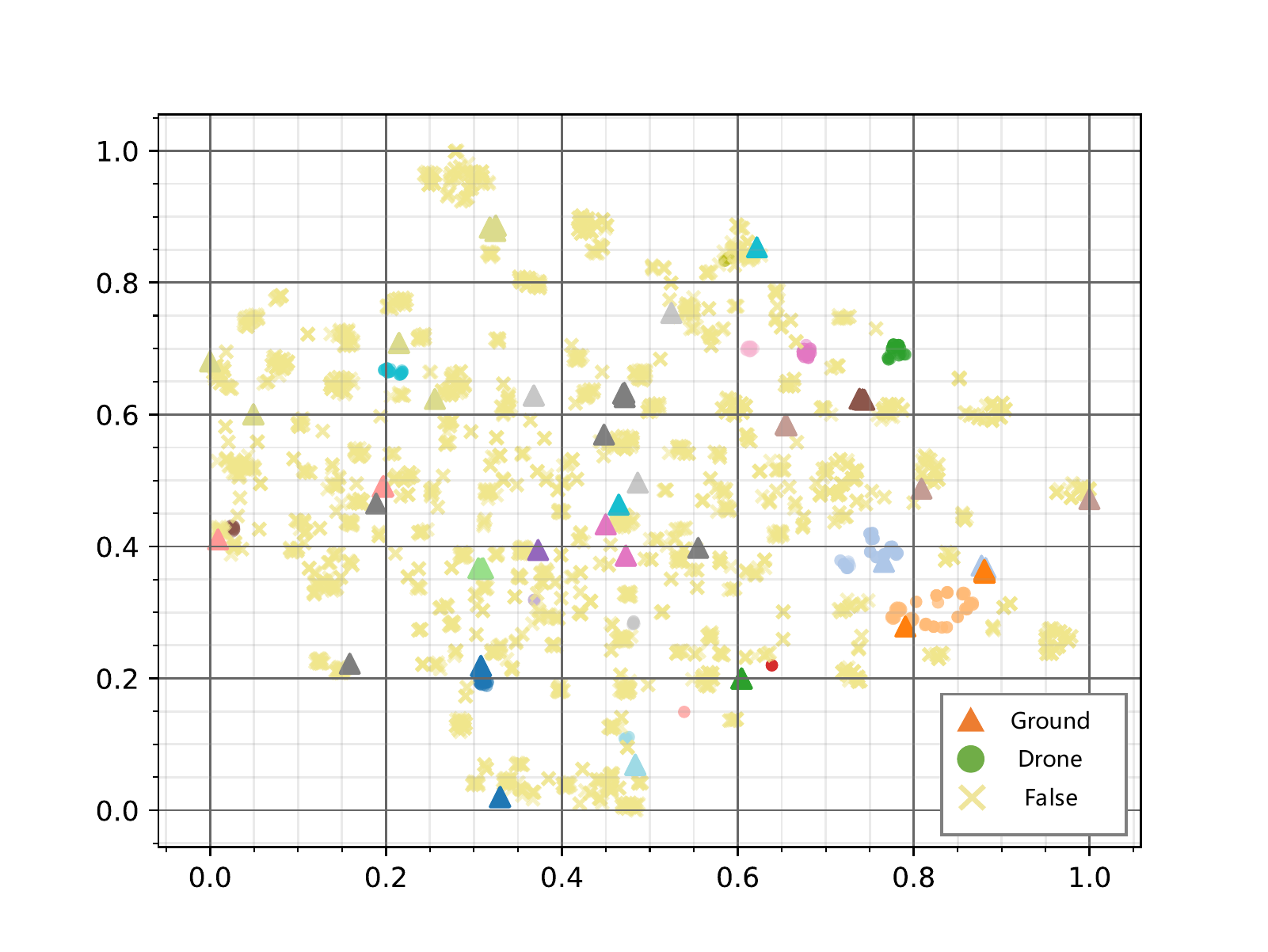} &
        \includegraphics[width=0.2\textwidth]{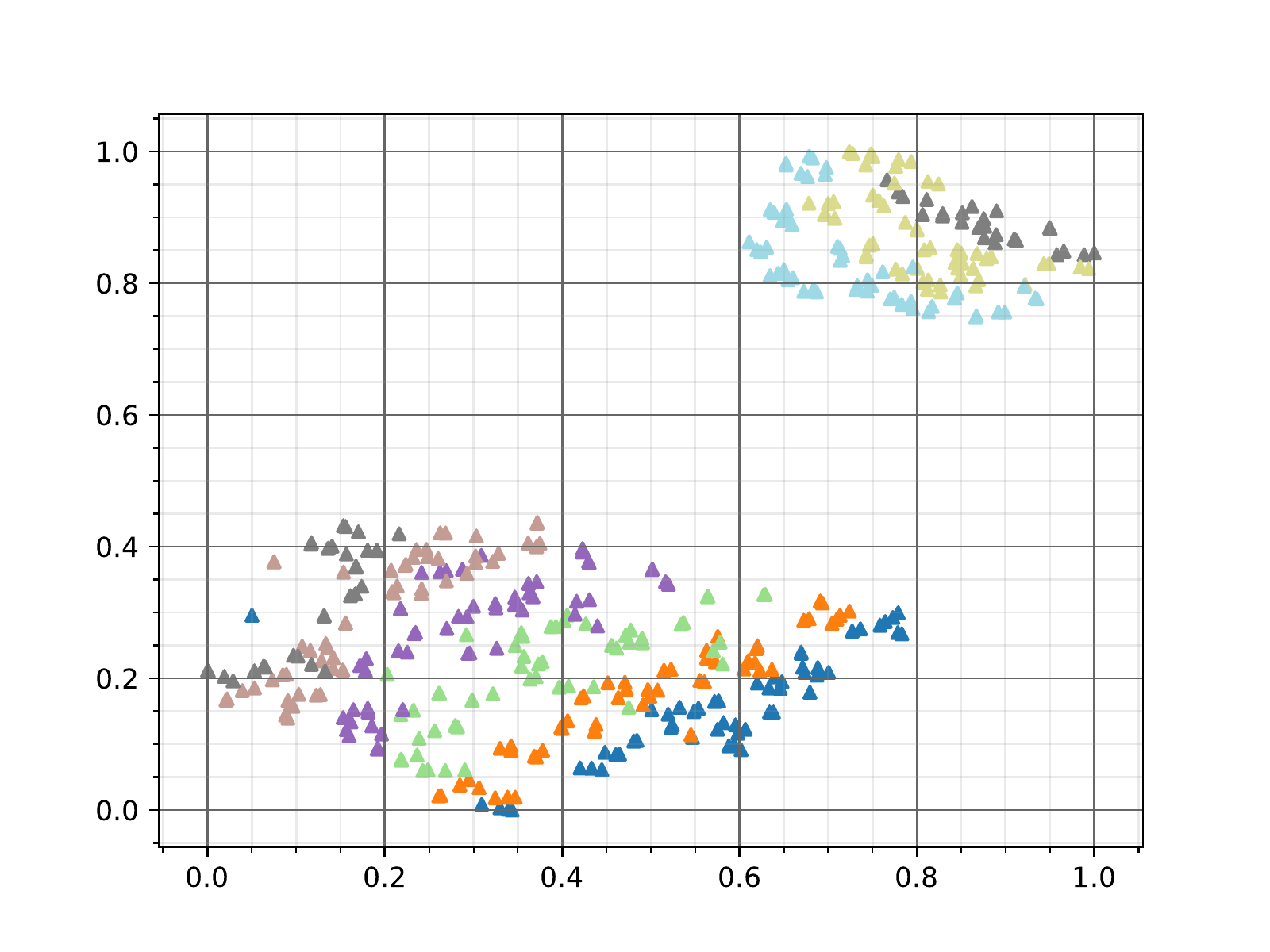} &
        \includegraphics[width=0.2\textwidth]{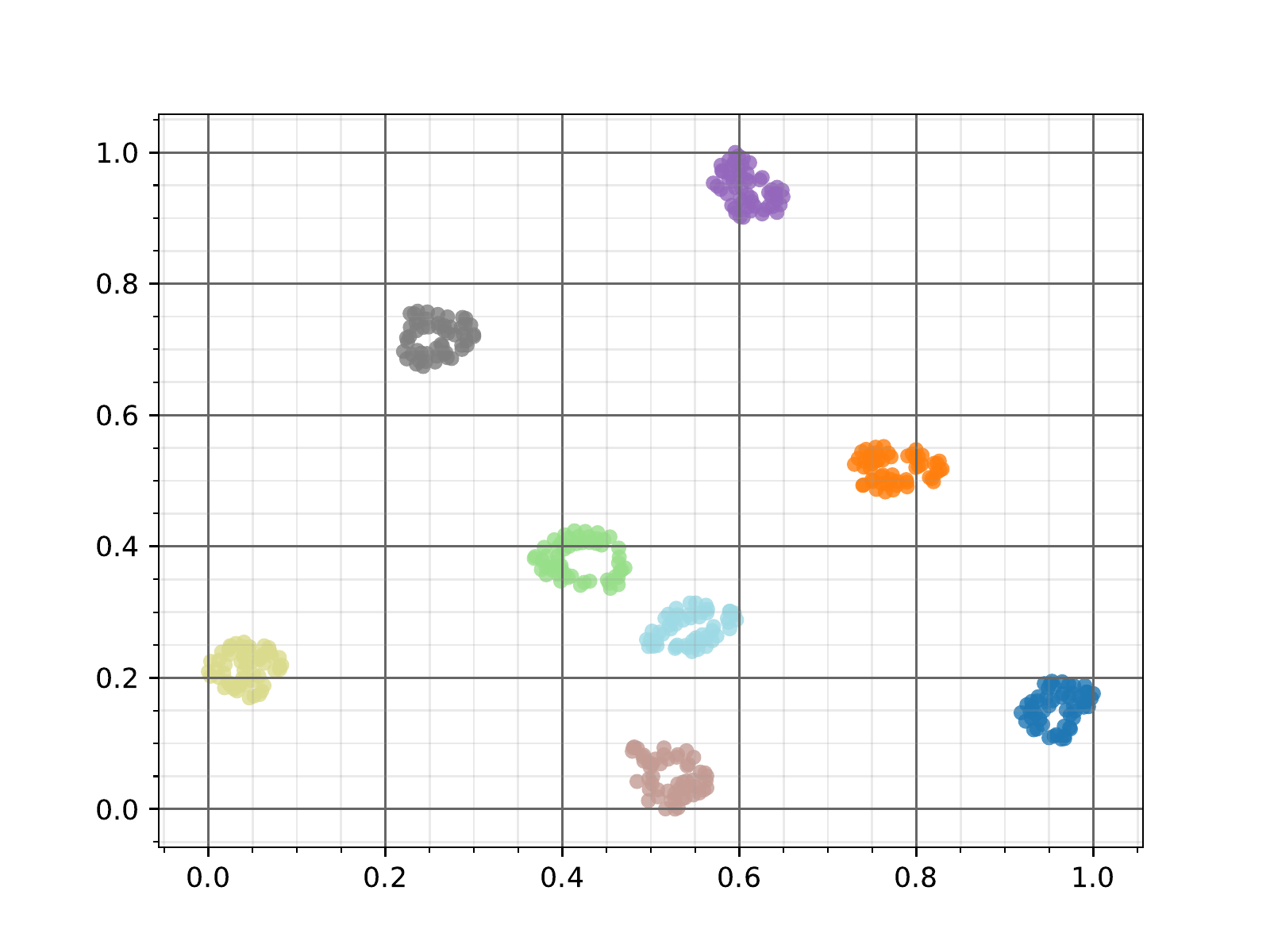} &
        \includegraphics[width=0.2\textwidth]{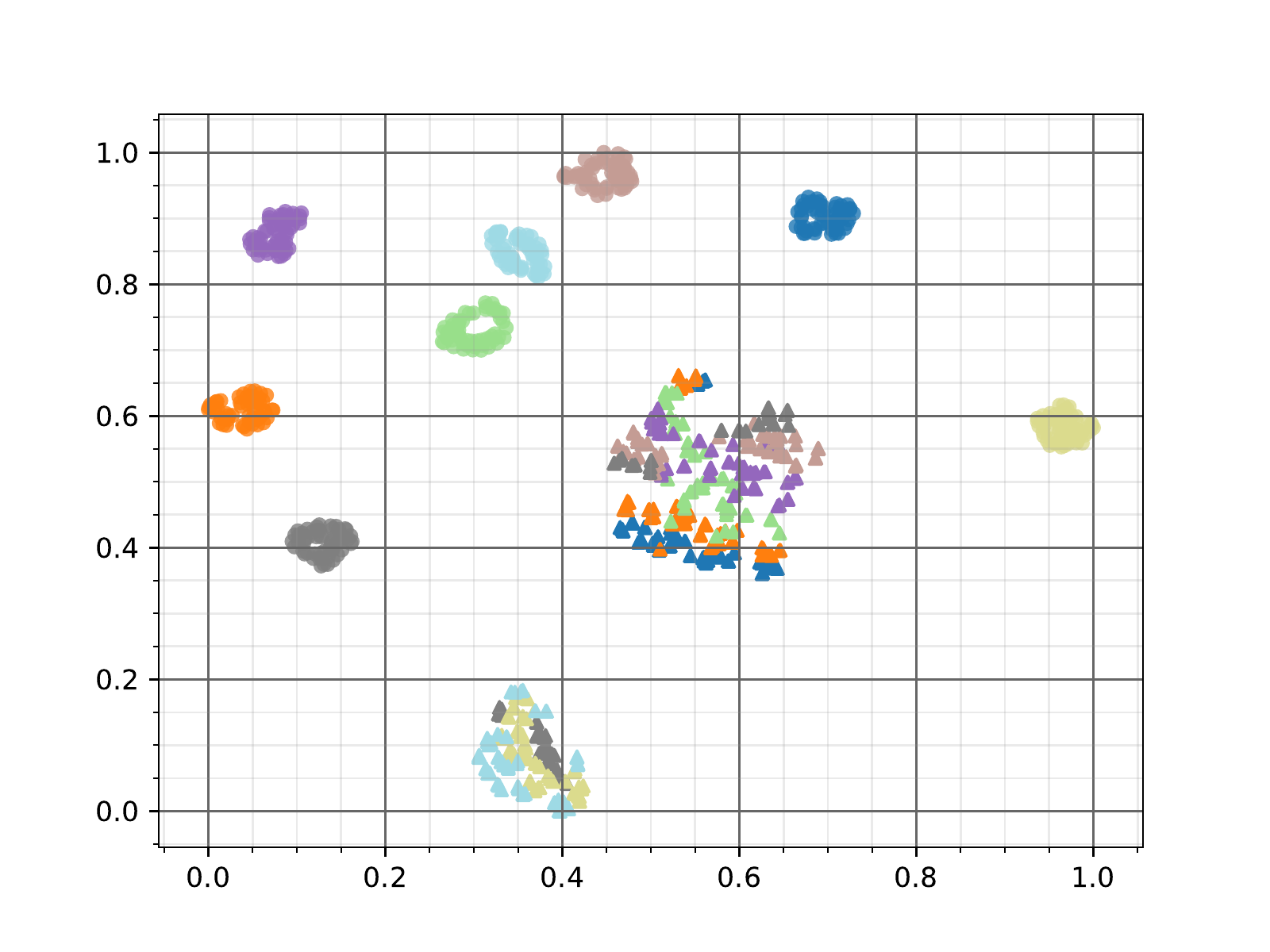}\\
        \small{(a)} & \small{(b)} & 
        \small{(c)} & \small{(d)} & \small{(e)}\\
    \end{tabular}
    \vspace{-3mm}
    \caption{Visualization of feature distributions. (a) and (b) show the feature distributions of peer learning and Str-CNN~\cite{zheng2020university}, respectively. For each ground-view image (donated by the triangle), we select its top-50 retrieval results from drone-view gallery list. Positive and negative results are donated as circles and $\times$ marks, respectively. (c), (d), (e) show the feature distributions of ground-drone, satellite-drone, and ground-drone-satellite spaces, respectively. Features from the ground-drone and satellite-drone are denoted as triangles and circles, respectively.}
    \vspace{-2mm}
    \label{fig:distribution}
\end{figure*}

We sample some drone-view images from University-Earth's testing set and visualize their feature distribution in ground-drone (peer learning model) and satellite-drone (patch-based model) spaces. Then we project all features into the same space called ground-drone-satellite space. Figure~\ref{fig:distribution} (c)-(e) show the results. We can find that, in the ground-drone and satellite-drone spaces, features of each identity are relatively grouped in clusters. But even we try to project drone-view features from these two spaces to the same space, most of the samples are off the manifolds of their corresponding classes. This supports our assumption that mapping images from different space into a uniform space distorts the manifolds.

Note that cross diffusion requires the creation of graphs in two independent spaces. The quality of the graphs affects the final diffusion result, and peer learning can help us build better mapping spaces and thus improve the quality of the graphs, so peer learning is not a redundant part. We also apply diffusion on Str-CNN + Multi-loss and CVM-Net, achieved 28.29\% CMC@1\% accuracy, 10.03\% mAP and 26.18\% CMC@1\% accuracy, 9.21\% mAP respectively on ground-drone-satellite image retrieval (on the University-Earth dataset), all less accurate than PLCD. Hence, Table.~\ref{tab:diffusion} also shows that simply building a connection between two independent spaces does not achieve good results and also illustrates the effectiveness of diffusion.

\subsection{How important are the Drone-view Images?}
To demonstrate that drone-view images can facilitate ground-to-satellite image retrieval task, we select state-of-the-art methods and train them on the ground and satellite sets, or on the ground, drone, and satellite sets. The results are shown in Table.~\ref{tab:sota}. In the table, notation with a `\textit{w/o}’ suffix means that the indicated model is only trained on the ground and satellite sets without drone-view images. The results show that every method can improve its performance by adding drone-view images for training. The performance demonstrates a fact that drone-view information benefits the ground-to-satellite image retrieval task, and plays an important role to make a bridge between ground-view and satellite-view images. 

In our method, drone images (\textbf{no label}) are set as the reference set at query time. The content overlap between ground and satellite images of the same landmark is very small, it is pretty difficult to extract relevant information between them without the drone images. And due to the diversity of landmarks, even if drone images are used in training, it is difficult for the model to make correct inferences at query time if drone images are not used just as a reference. 

 Here we would like to discuss the practicality of using drone data as a reference. In the real world, it would be really difficult to collect drone images in some cities, but we can use synthetic data to solve this problem. In the dataset we used, all the ground images and satellite images are collected in the real world, while the drone images are \textbf{synthetic data} collected from google-earth. From the experiments, we can see that the synthetic drone images help to improve the ground-stallite retrieval task. In addition, in the test, the drone data as reference does not need any annotation, so in the real application, we only need to provide the model with the synthetic data of the corresponding area. We believe our task fits the realistic application well.

\subsection{Why do not use One Model for All View?}

The viewpoint varies greatly between ground-view and satellite-view/drone-view images. We consider it's hard to design a common representation model for images of all views. To prove this, we make a comparison of one common model and the two-branch model in the setting of Ground$\rightarrow$Drone retrieval, Drone$\rightarrow$Satellite retrieval and Ground$\rightarrow$Satellite retrieval. Table.~\ref{tab:one common model} shows the results of all settings trained on the University-Earth dataset. The results clearly demonstrate that the two-branch structure outperforms the one common model in all settings. Thus, a two-branch structure is more suitable for this kind of ground-drone-satellite retrieval tasks that have large viewpoint changes.

\subsection{Experiments on the parameters}
The $\tau$ is the temperature hyper-parameter. When the value of $\tau$ less than 1, it makes the similarity vector sharper and vice versa. We do not expect that the distribution space of the junior model is too similar to the senior model. We consider that making the similarity vector sharper can improve the power of supervision from the senior model and achieves better results. 

\begin{figure}[t]
\centering
\begin{tabular}{cc}
    \includegraphics[width=0.49\linewidth]{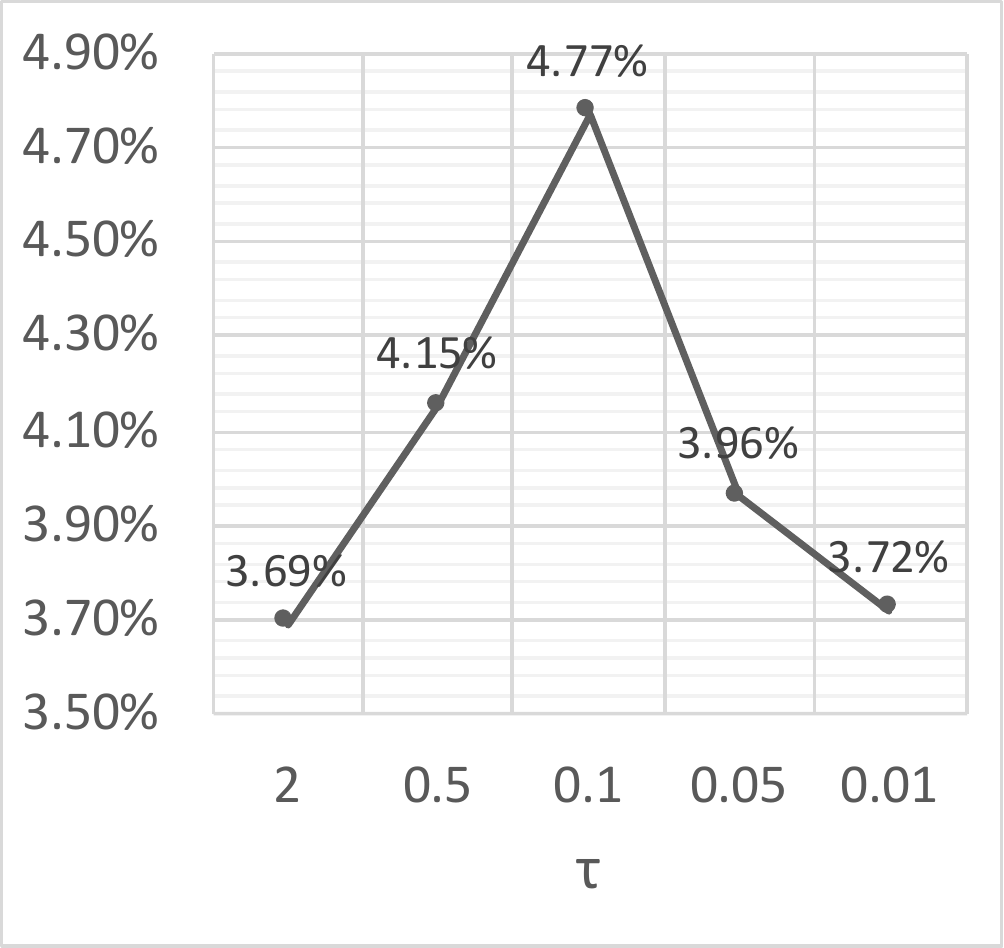} &
		\includegraphics[width=0.49\linewidth]{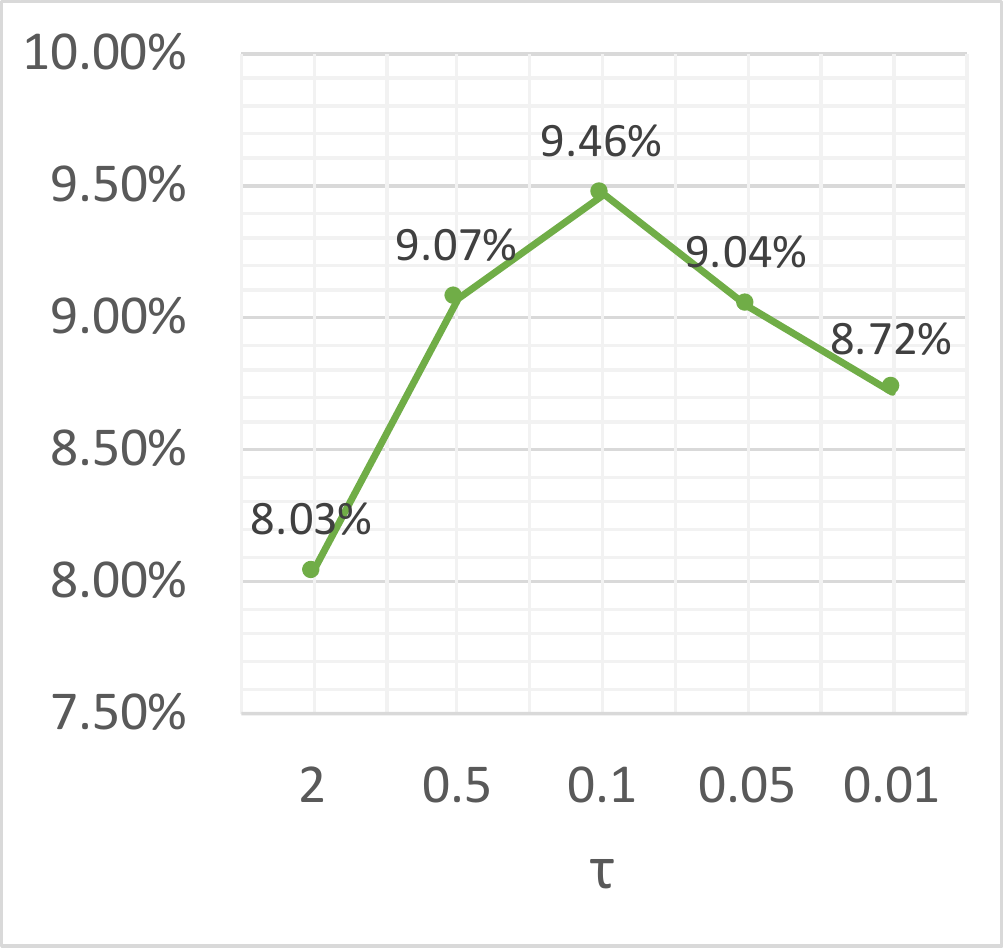} \\
		\footnotesize{(a) CMC@1 Results} & \footnotesize{(b) CMC@5 Results}\\
		\includegraphics[width=0.49\linewidth]{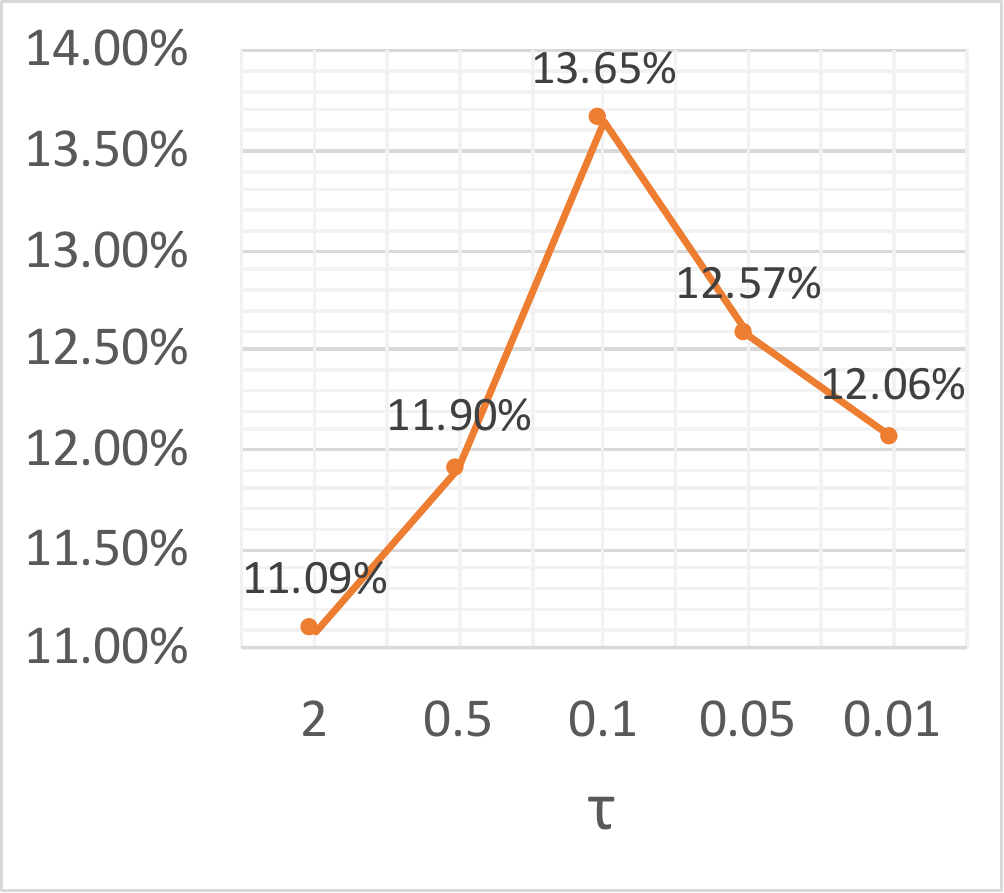} &
		\includegraphics[width=0.49\linewidth]{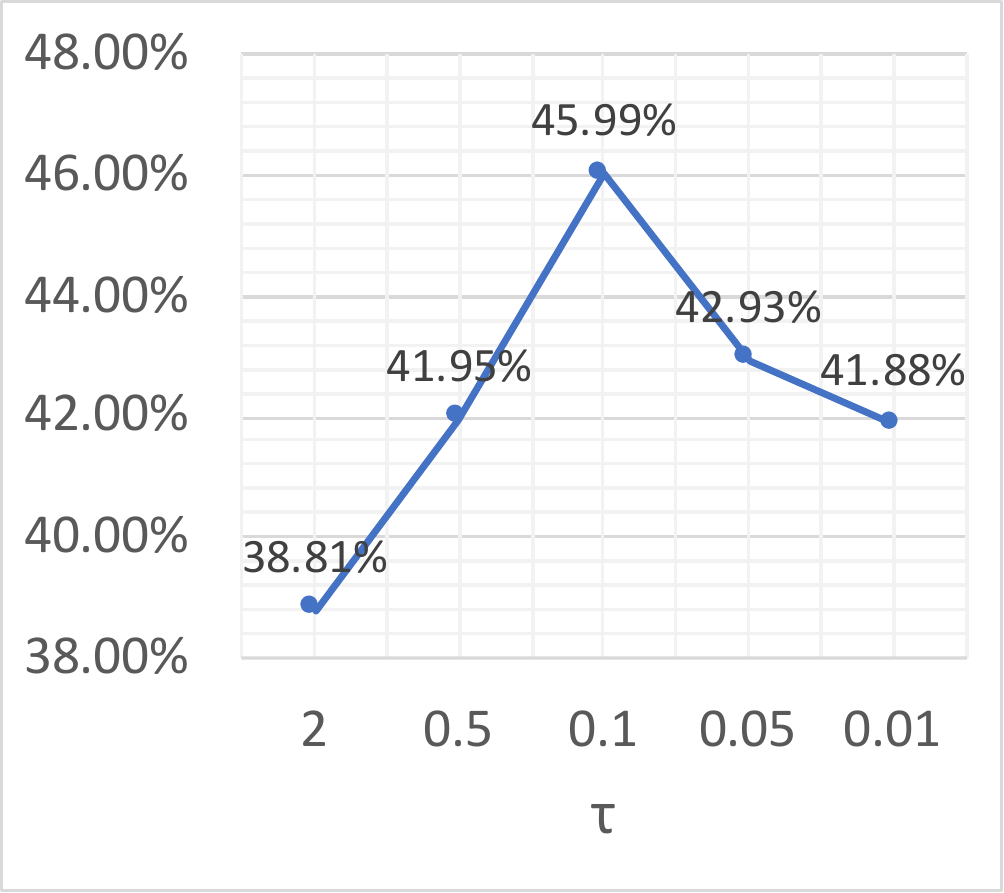} \\
		\footnotesize{(a) CMC@10 Results} & \footnotesize{(b) CMC@1\% Results}\\
\end{tabular}  
\vspace{-3mm}
  \caption{Results on different parameter $\tau$.}
  \label{fig:tau}
\end{figure}

To support our assumption, we conducted experiments with different values of $\tau$ ($\tau=2,0.5,0.1,0.05$ and $0.01$) on the University-Earth dataset. Figure~\ref{fig:tau} reports the results. The results show that our model achieves the best result when $\tau=0.1$. 

{We also have experiments on the parameters $\lambda_{1}$ and $\lambda_{2}$. The results are shown in Table~\ref{tab:parameters}, which indicate that our model achieves the best result when $\lambda_{1} = 1$ and $\lambda_{2} = 1$. }

\tabcolsep=1pt

\begin{figure}[t]
\centering
    \includegraphics[width=1\linewidth]{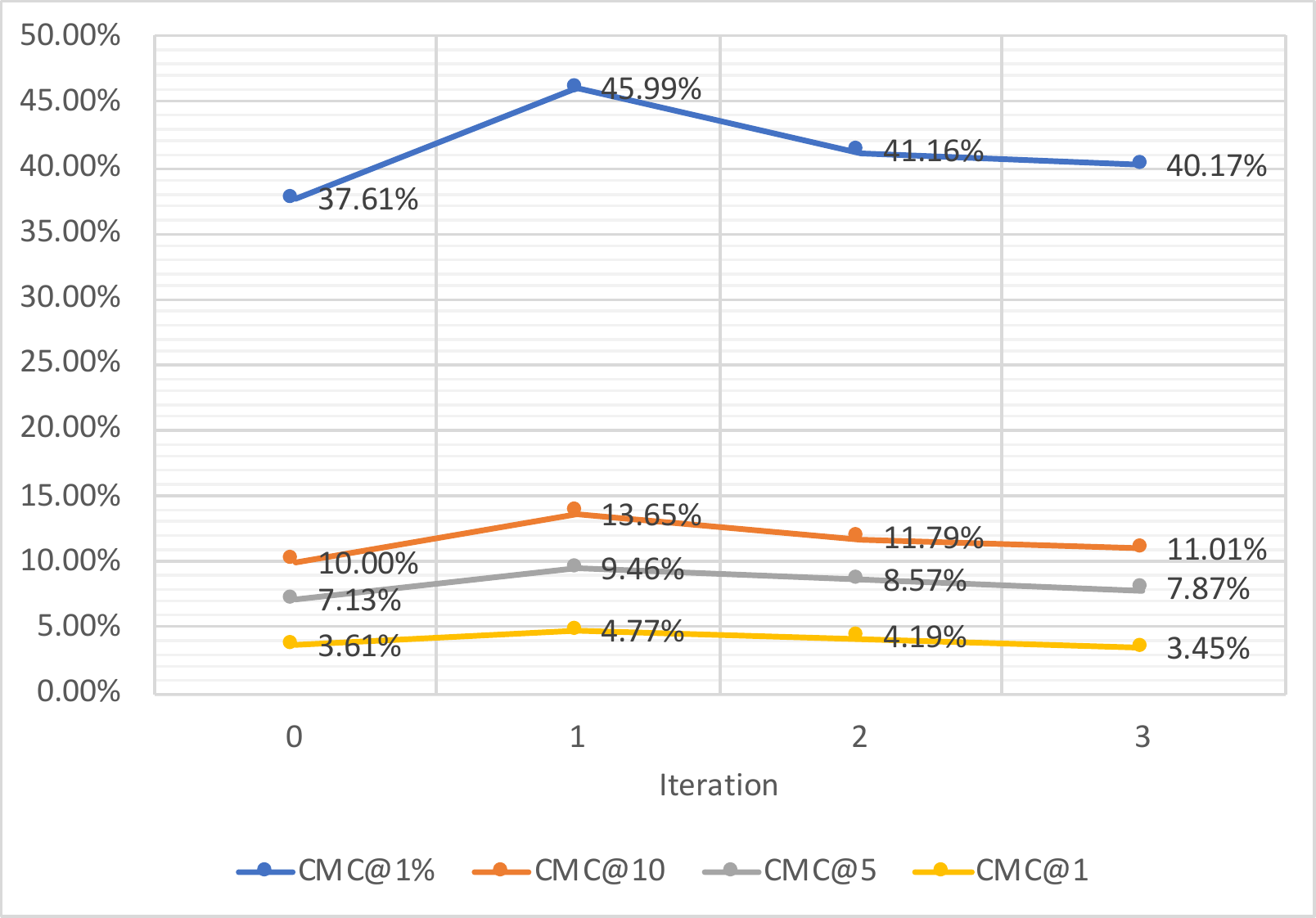}
  \caption{Results on different iterations ($\tau = 0.1$).}
  \label{fig:generation}
\end{figure}

\begin{table}[!h]
    \centering
    \tabcolsep=12pt
    \begin{tabular}{c|c|c}
    \hline
          $\lambda_1$ & CMC@1\% & mAP (\%)\\\hline
       0.1&  42.65&  5.36\\
       0.5&  43.62&  6.05\\
       $\boldsymbol{1}$  &  $\boldsymbol{45.99}$& $\boldsymbol{6.32}$\\
       2  &  41.99& 5.86 \\
       5  &  33.89& 4.12 \\
      \hline
    \end{tabular}
    \hspace{5mm}
    \begin{tabular}{c|c|c}
    \hline
          $\lambda_2$ & CMC@1\% & mAP (\%)\\\hline
       0.1 &  80.18&  64.64\\
       0.5&  86.10&  62.25\\
       $\boldsymbol{1}$  &  $\boldsymbol{87.69}$& $\boldsymbol{70.60}$\\
       2  &  83.72& 63.37 \\
       5  &  79.89& 54.12 \\
      \hline
    \end{tabular}
    \caption{Results on different parameters $\lambda_1$ and $\lambda_2$.}
    \label{tab:parameters}
\end{table}

{
\subsection{The number of training samples}
We conducted experiments on the University-Earth dataset by using different batchsize. The batchsize $N$ means that we randomly select $N$ street/satellite images and $6N$ drone images in each mini-batch. The results of different batchsize are shown in Table~\ref{tab:batchsize}. Hence, considering the batchsize for the best performance, we set the batchsize to 8/4 in our work.
}
\begin{table}[!h]
    \centering
    \tabcolsep=12pt
    \begin{tabular}{c|c|c}
    \hline
          $N$ & CMC@1\% & mAP (\%)\\\hline
       4&  30.24&  2.77\\
       6&  39.32&  4.78\\
       $\boldsymbol{8}$  &  $\boldsymbol{45.99}$& $\boldsymbol{6.32}$\\
       12 (two GPUs)  &  41.84& 4.76 \\
       16 (two GPUs) &  41.57& 4.49 \\
      \hline
    \end{tabular}
    \hspace{5mm}
    \begin{tabular}{c|c|c}
    \hline
          $N$ & CMC@1\% & mAP (\%)\\\hline
       2&  71.71&  52.54\\
       $\boldsymbol{4}$ & $\boldsymbol{87.69}$ &  $\boldsymbol{70.60}$\\
       6  &  86.51& 68.47\\
       8  &  84.17& 63.91 \\
       10  &  80.73& 59.83 \\
      \hline
    \end{tabular}
    \caption{Results on different batchsize $N$ for Ground-Drone Representation and Drone-satellite Representation, respectively.}
    \label{tab:batchsize}
\end{table}

\subsection{Experiments on the iteration of peer learning}
Soft label estimation has been widely studied in self-supervised learning methods, where the network predicts soft labels by properly utilizing the capability of the network itself. Generally speaking, the model is trained in several runs via self-predicted soft labels. Some previous works~\cite{furlanello2018born,xie2020self} demonstrate the effectiveness of progressive learning in iteration. 

To investigate the effectiveness of peer learning in iteration via soft labels, we conducted experiments on the University-Earth dataset with different iterations. In each iteration, we use the junior model from the last iteration as the senior model to finish the peer learning process. We set $\tau = 0.1$. Figure~\ref{fig:generation} shows the results. The results show that our model achieves the best result in the first iteration.

\tabcolsep=1pt
\begin{figure*}[t]
	\centering
		\begin{tabular}{l|lllll|lllll}
		\small{Query} & \multicolumn{5}{l|}{Ground $\rightarrow$ Drone (Top1 $\rightarrow$ Top5)}
		& \multicolumn{5}{l}{Ground $\rightarrow$ Satellite (Top1 $\rightarrow$ Top5)}\\
		\includegraphics[width=0.09\linewidth]{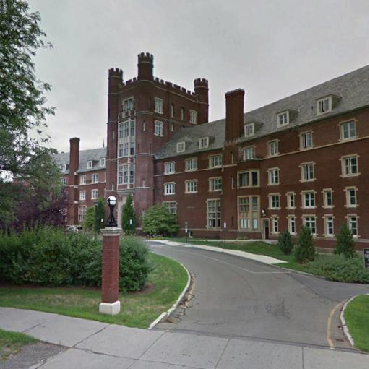} &
		\includegraphics[width=0.09\linewidth]{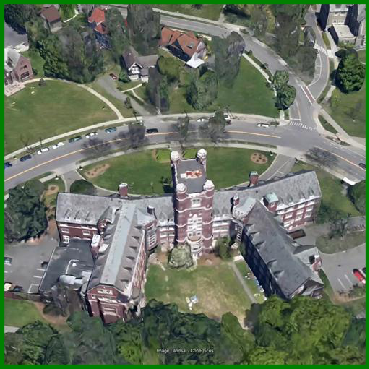} &
		\includegraphics[width=0.09\linewidth]{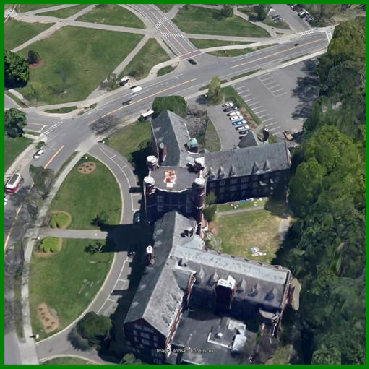} &
		\includegraphics[width=0.09\linewidth]{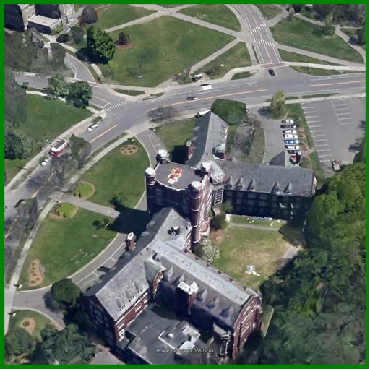} &
		\includegraphics[width=0.09\linewidth]{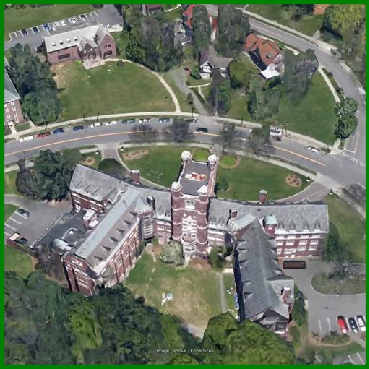} &
		\includegraphics[width=0.09\linewidth]{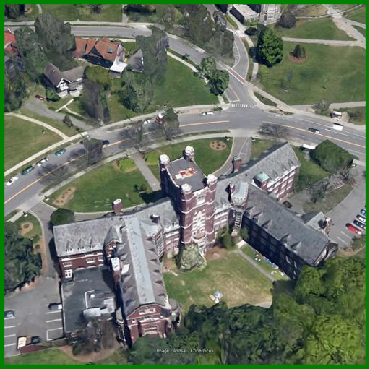} &
		\includegraphics[width=0.09\linewidth]{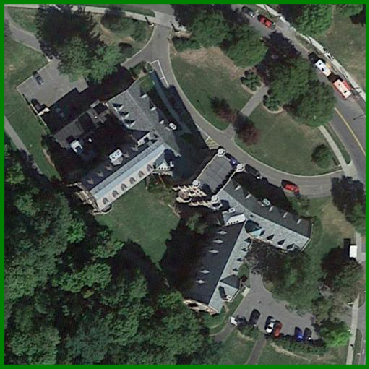} &
		\includegraphics[width=0.09\linewidth]{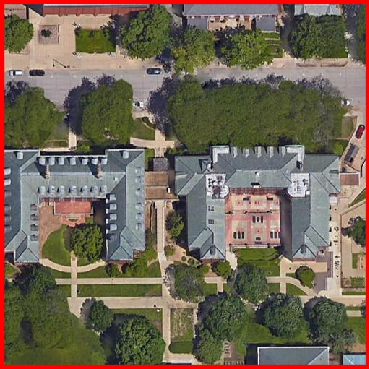} &
		\includegraphics[width=0.09\linewidth]{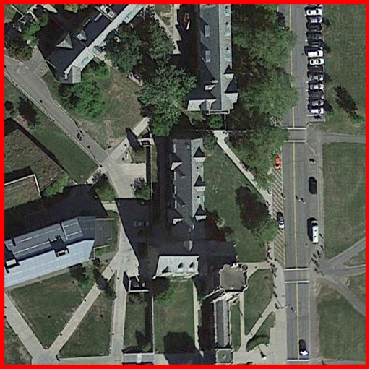} &
		\includegraphics[width=0.09\linewidth]{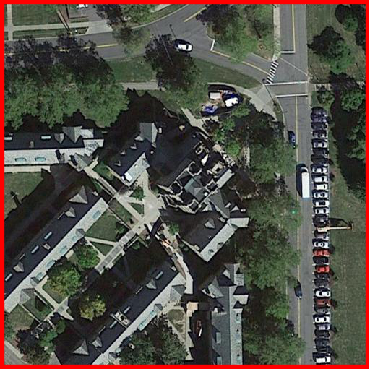} &
		\includegraphics[width=0.09\linewidth]{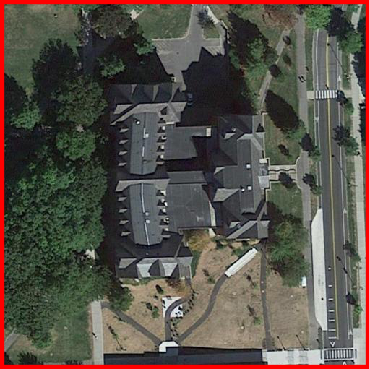} \\
		\includegraphics[width=0.09\linewidth]{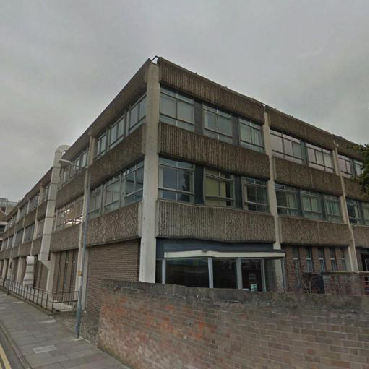} &
		\includegraphics[width=0.09\linewidth]{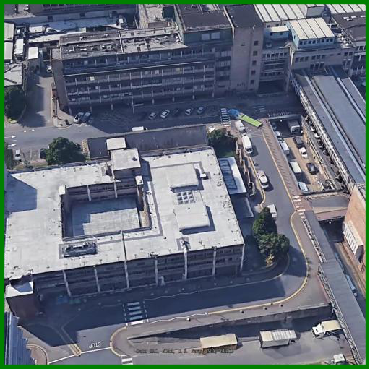} &
		\includegraphics[width=0.09\linewidth]{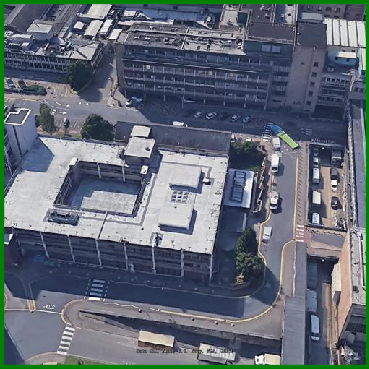} &
		\includegraphics[width=0.09\linewidth]{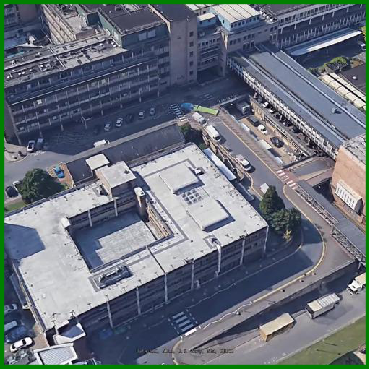} &
		\includegraphics[width=0.09\linewidth]{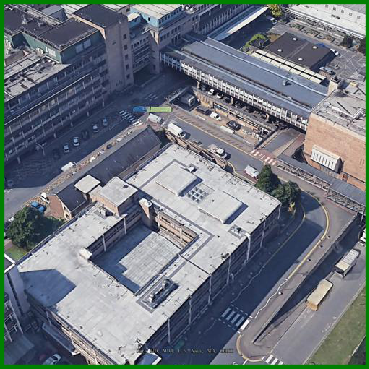} &
		\includegraphics[width=0.09\linewidth]{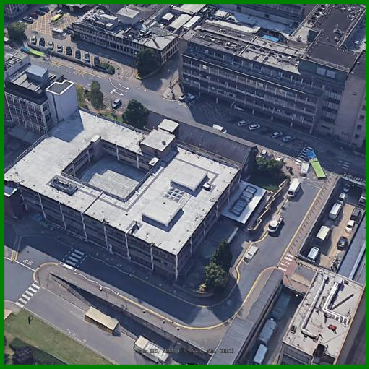} &
		\includegraphics[width=0.09\linewidth]{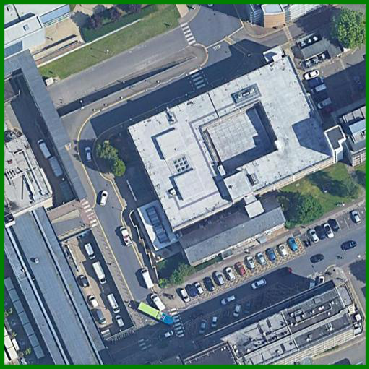} &
		\includegraphics[width=0.09\linewidth]{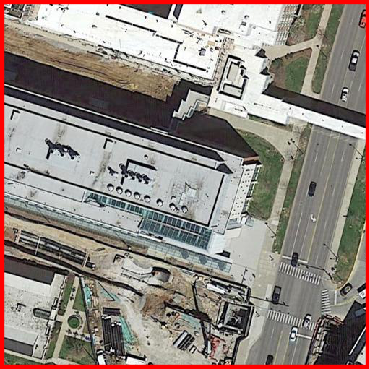} &
		\includegraphics[width=0.09\linewidth]{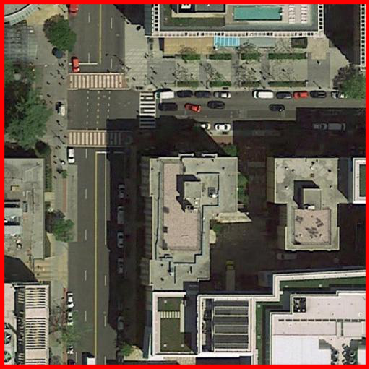} &
		\includegraphics[width=0.09\linewidth]{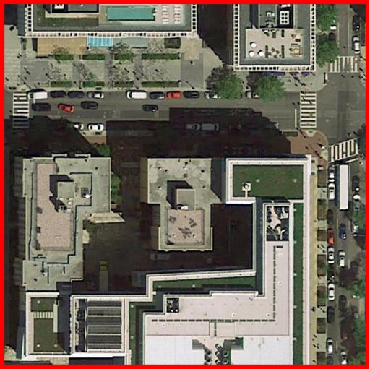} &
		\includegraphics[width=0.09\linewidth]{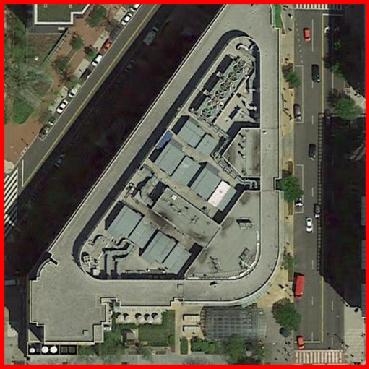} \\	
		\includegraphics[width=0.09\linewidth]{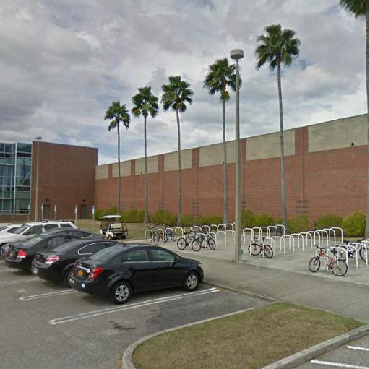} &
		\includegraphics[width=0.09\linewidth]{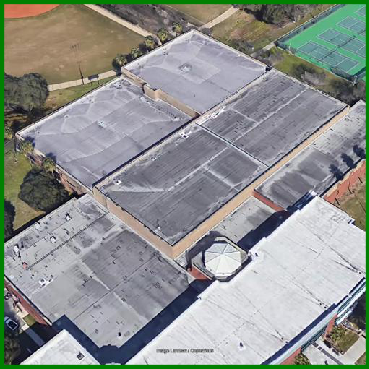} &
		\includegraphics[width=0.09\linewidth]{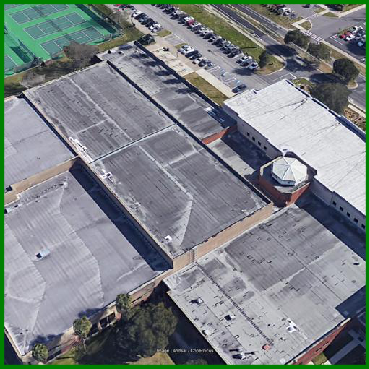} &
		\includegraphics[width=0.09\linewidth]{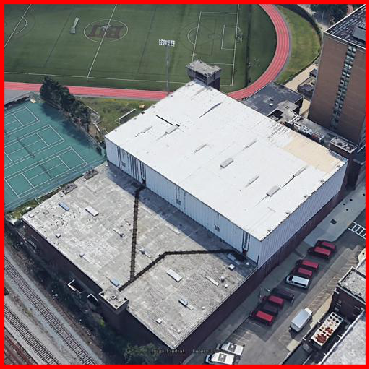} &
		\includegraphics[width=0.09\linewidth]{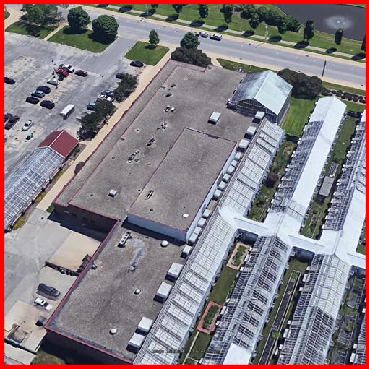} &
		\includegraphics[width=0.09\linewidth]{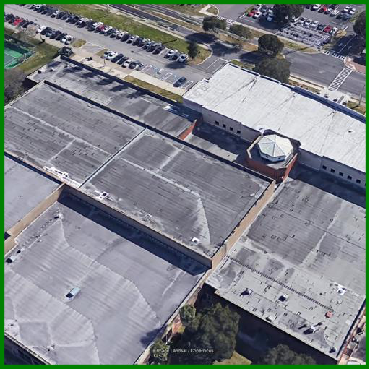} &
		\includegraphics[width=0.09\linewidth]{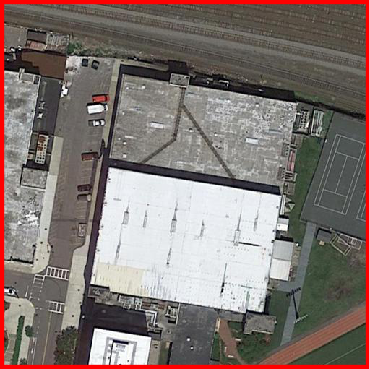} &
		\includegraphics[width=0.09\linewidth]{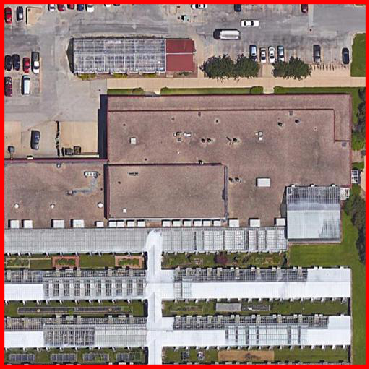} &
		\includegraphics[width=0.09\linewidth]{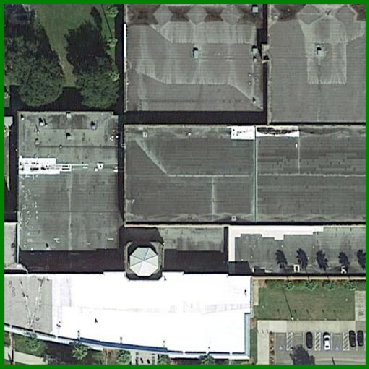} &
		\includegraphics[width=0.09\linewidth]{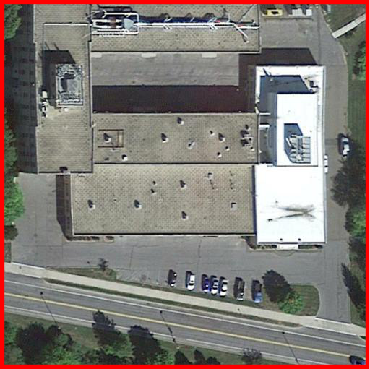} &
		\includegraphics[width=0.09\linewidth]{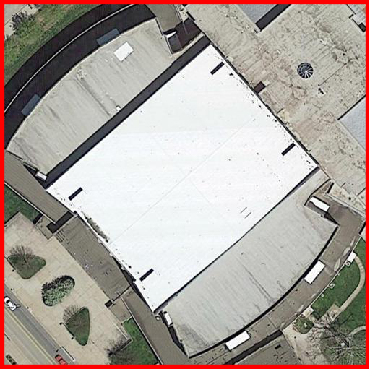} \\
		\includegraphics[width=0.09\linewidth]{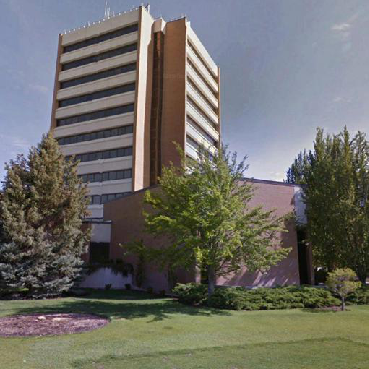} &
		\includegraphics[width=0.09\linewidth]{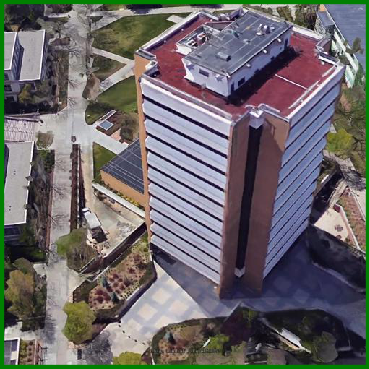} &
		\includegraphics[width=0.09\linewidth]{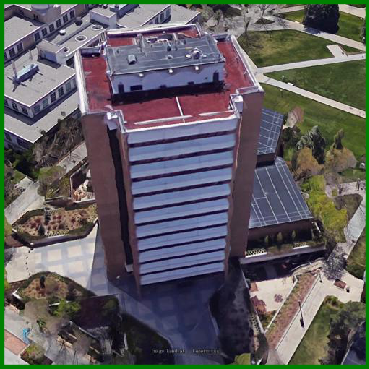} &
		\includegraphics[width=0.09\linewidth]{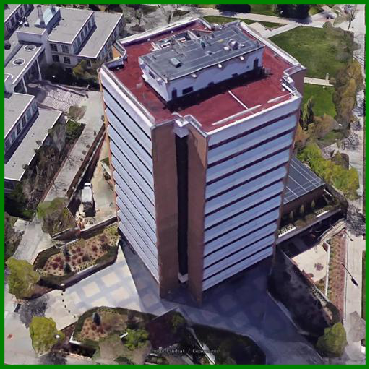} &
		\includegraphics[width=0.09\linewidth]{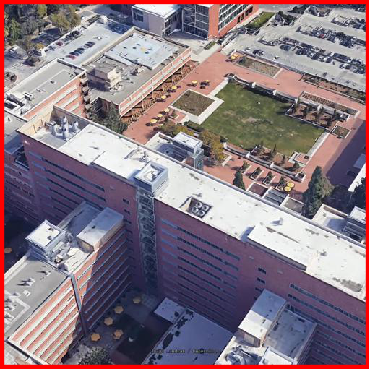} &
		\includegraphics[width=0.09\linewidth]{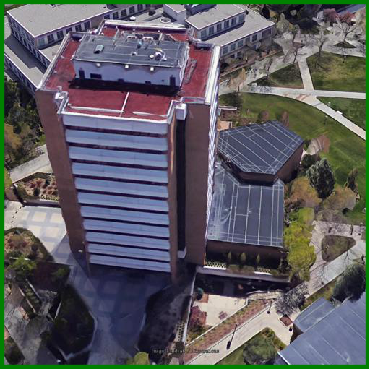} &
		\includegraphics[width=0.09\linewidth]{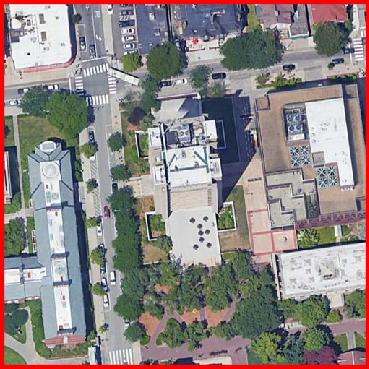} &
		\includegraphics[width=0.09\linewidth]{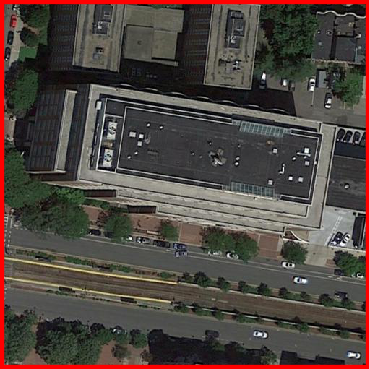} &
		\includegraphics[width=0.09\linewidth]{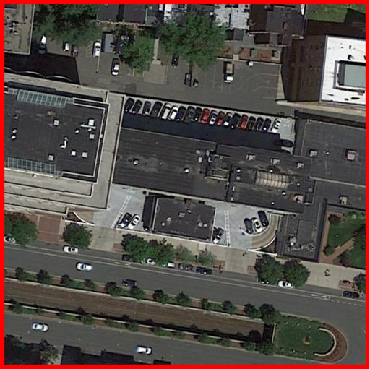} &
		\includegraphics[width=0.09\linewidth]{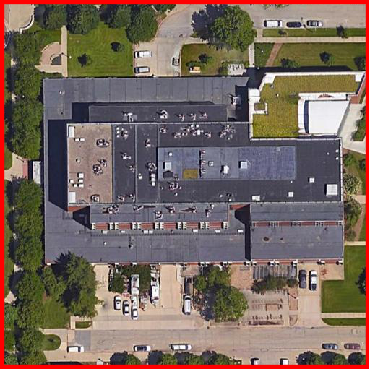} &
		\includegraphics[width=0.09\linewidth]{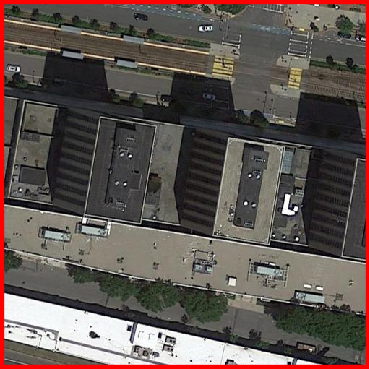} \\
		\includegraphics[width=0.09\linewidth]{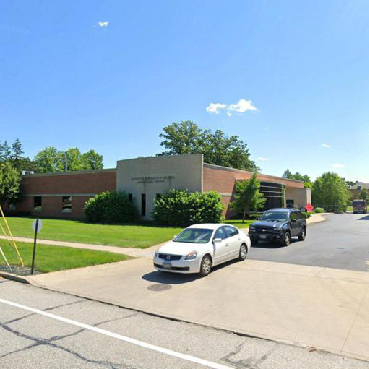} &
		\includegraphics[width=0.09\linewidth]{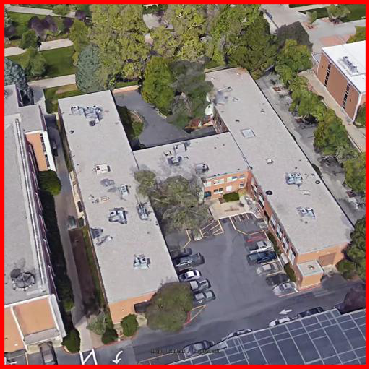} &
		\includegraphics[width=0.09\linewidth]{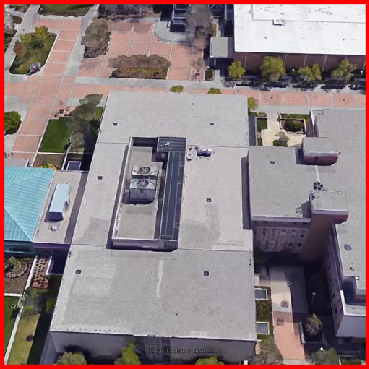} &
		\includegraphics[width=0.09\linewidth]{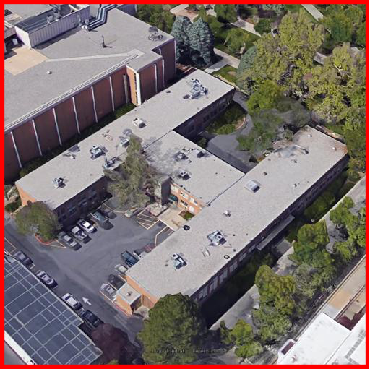} &
		\includegraphics[width=0.09\linewidth]{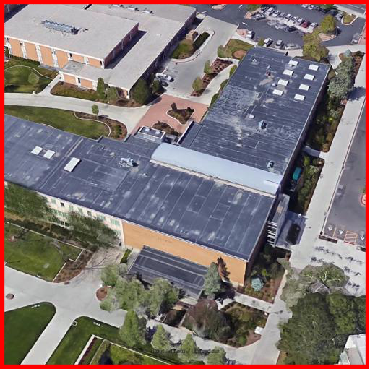} &
		\includegraphics[width=0.09\linewidth]{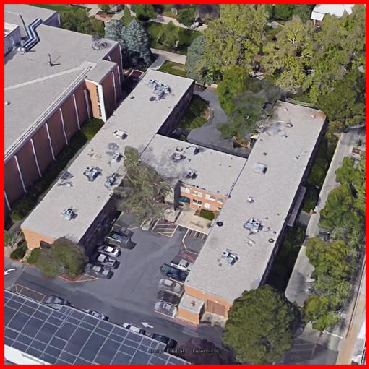} &
		\includegraphics[width=0.09\linewidth]{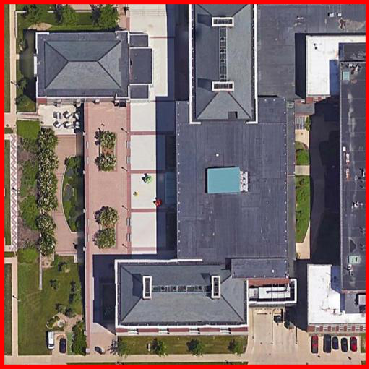} &
		\includegraphics[width=0.09\linewidth]{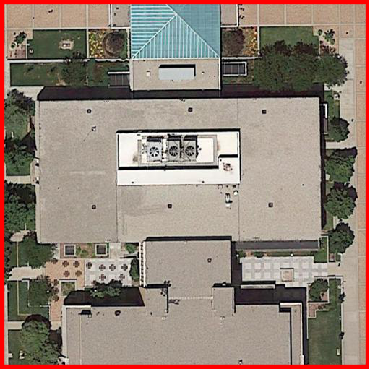} &
		\includegraphics[width=0.09\linewidth]{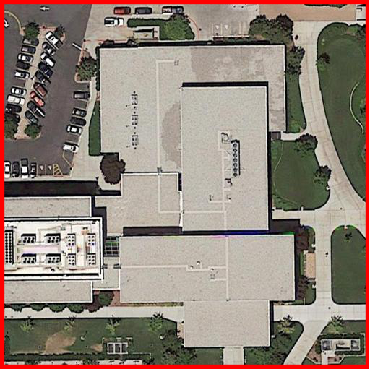} &
		\includegraphics[width=0.09\linewidth]{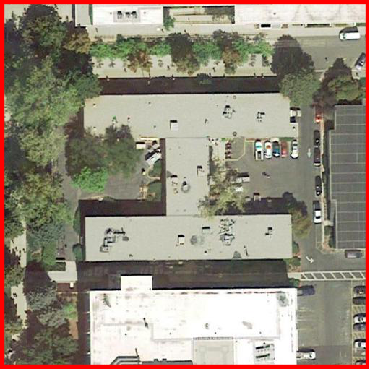} &
		\includegraphics[width=0.09\linewidth]{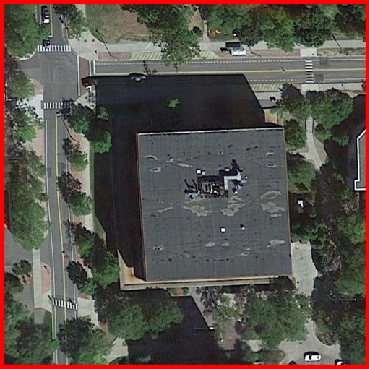} \\
       \end{tabular}
   \caption{The visualization of ground-to-satellite image retrieval by PLCD framework on University-Earth dataset. There are two results, from left to right: ground-view query image, the Top 1-5 retrieved drone-view images and the Top 1-5 retrieved satellite-view images. Green and red borders indicate correct and incorrect retrieved results, respectively.} 
\label{fig:result}
\end{figure*}

\tabcolsep=1pt
\begin{figure*}[t]
	\centering
		\begin{tabular}{l|lllll|lllll}
		\small{Query} & \multicolumn{5}{l|}{Ground $\rightarrow$ Drone (Top1 $\rightarrow$ Top5)}
		& \multicolumn{5}{l}{Ground $\rightarrow$ Satellite (Top1 $\rightarrow$ Top5)}\\
		\includegraphics[width=0.09\linewidth]{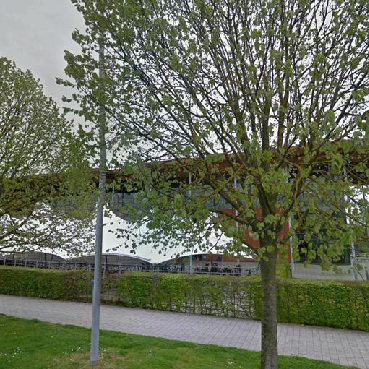} &
		\includegraphics[width=0.09\linewidth]{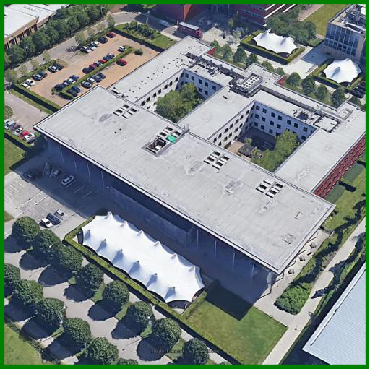} &
		\includegraphics[width=0.09\linewidth]{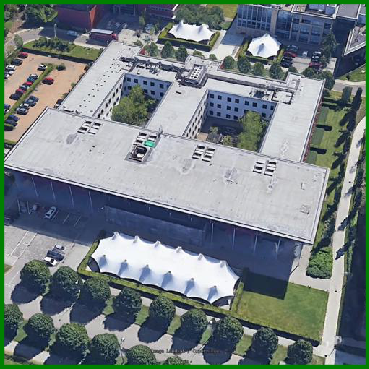} &
		\includegraphics[width=0.09\linewidth]{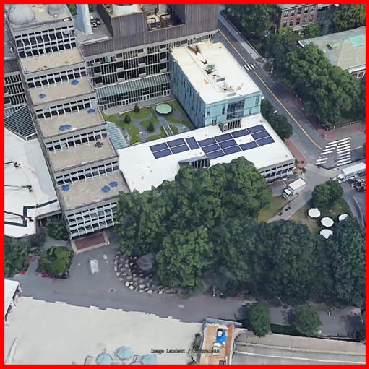} &
		\includegraphics[width=0.09\linewidth]{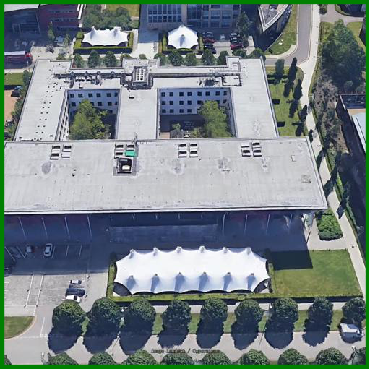} &
		\includegraphics[width=0.09\linewidth]{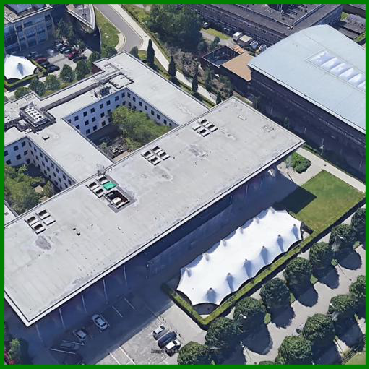} &
		\includegraphics[width=0.09\linewidth]{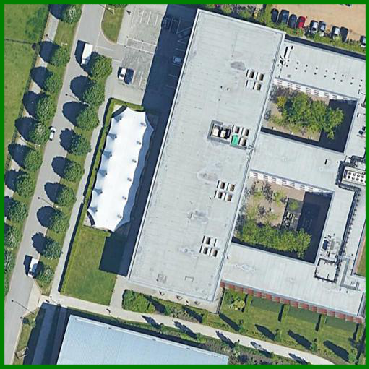} &
		\includegraphics[width=0.09\linewidth]{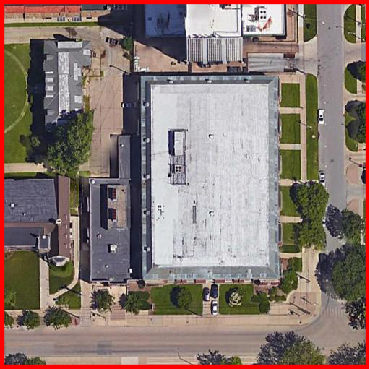} &
		\includegraphics[width=0.09\linewidth]{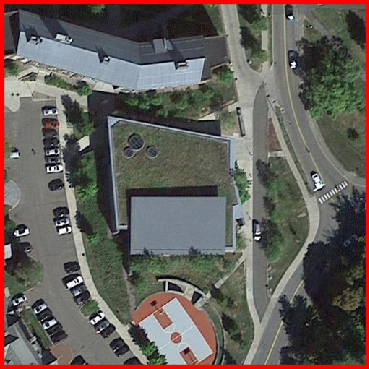} &
		\includegraphics[width=0.09\linewidth]{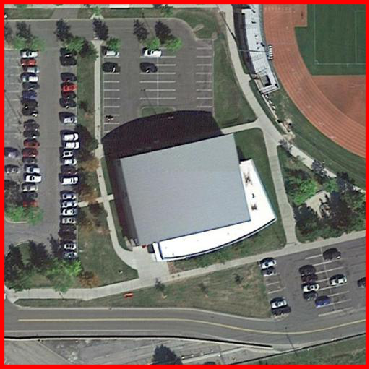} &
		\includegraphics[width=0.09\linewidth]{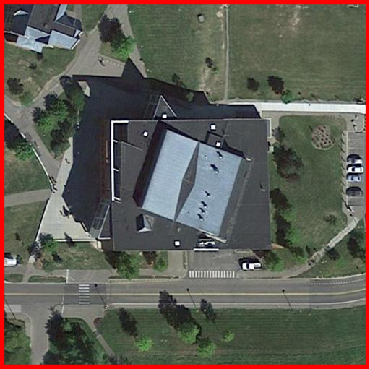} \\
		\includegraphics[width=0.09\linewidth]{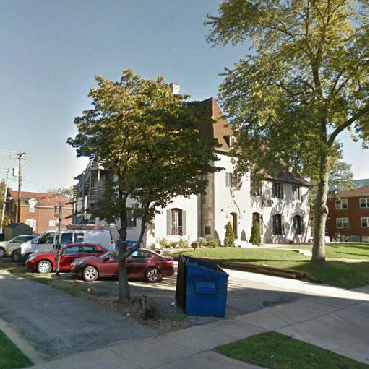} &
		\includegraphics[width=0.09\linewidth]{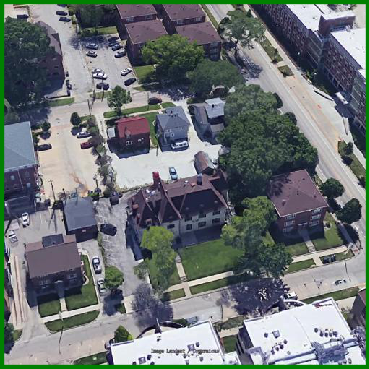} &
		\includegraphics[width=0.09\linewidth]{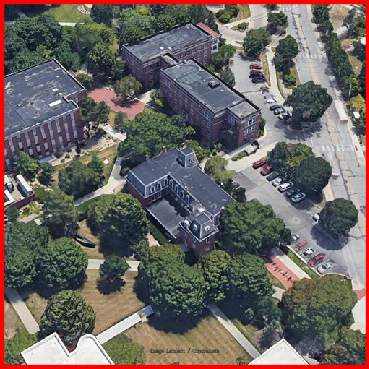} &
		\includegraphics[width=0.09\linewidth]{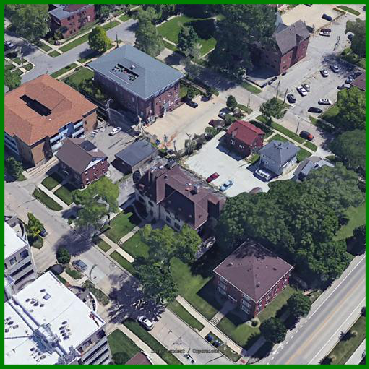} &
		\includegraphics[width=0.09\linewidth]{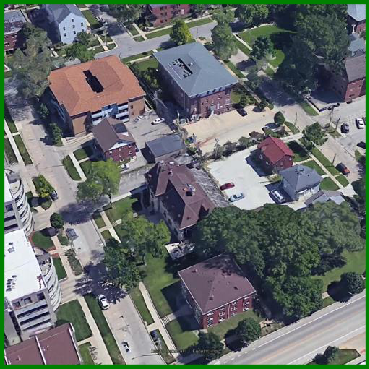} &
		\includegraphics[width=0.09\linewidth]{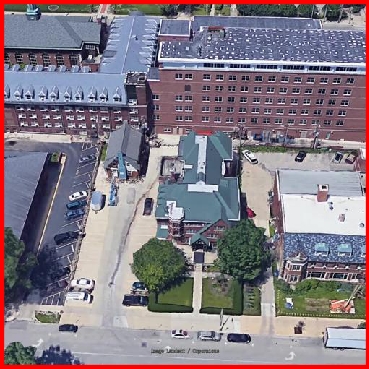} &
		\includegraphics[width=0.09\linewidth]{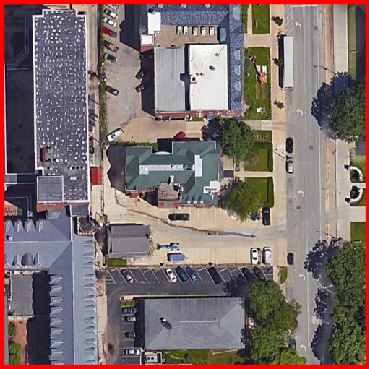} &
		\includegraphics[width=0.09\linewidth]{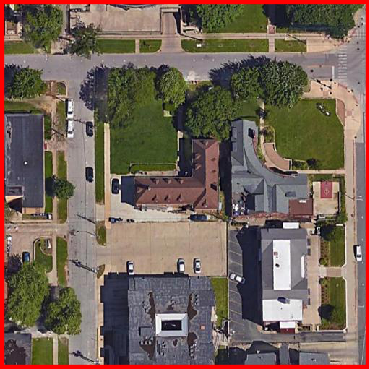} &
		\includegraphics[width=0.09\linewidth]{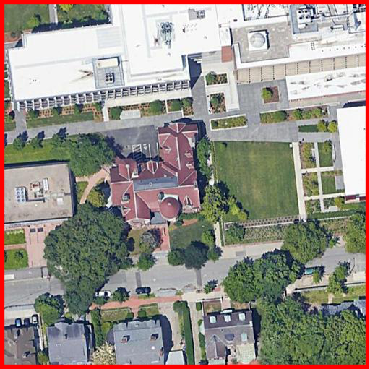} &
		\includegraphics[width=0.09\linewidth]{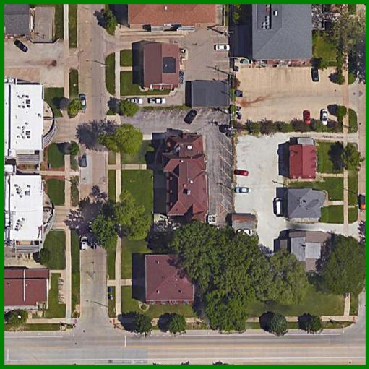} &
		\includegraphics[width=0.09\linewidth]{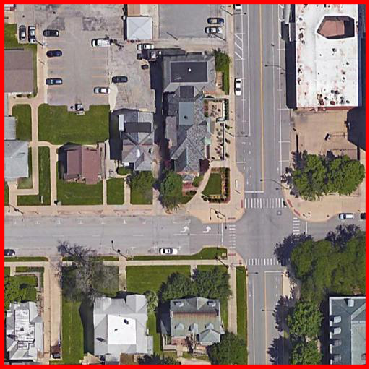} \\\includegraphics[width=0.09\linewidth]{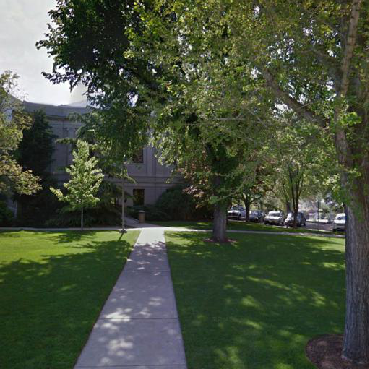} &
		\includegraphics[width=0.09\linewidth]{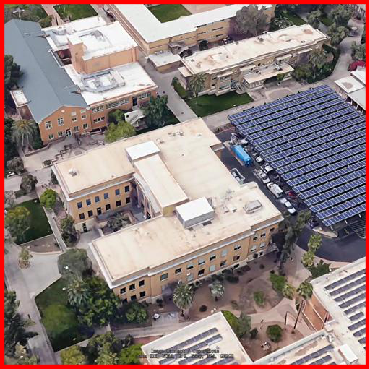} &
		\includegraphics[width=0.09\linewidth]{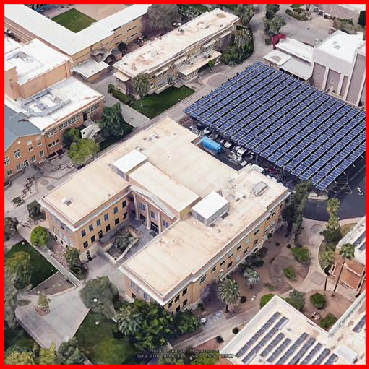} &
		\includegraphics[width=0.09\linewidth]{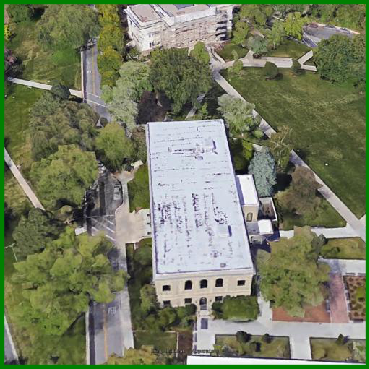} &
		\includegraphics[width=0.09\linewidth]{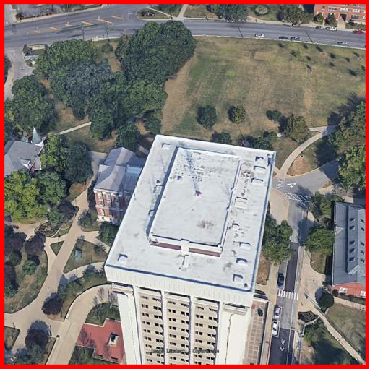} &
		\includegraphics[width=0.09\linewidth]{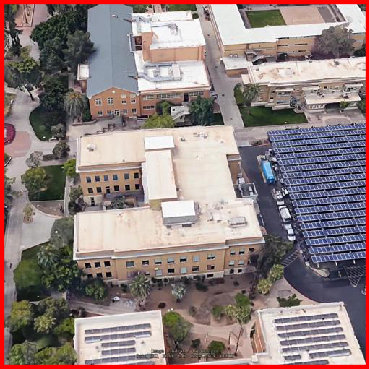} &
		\includegraphics[width=0.09\linewidth]{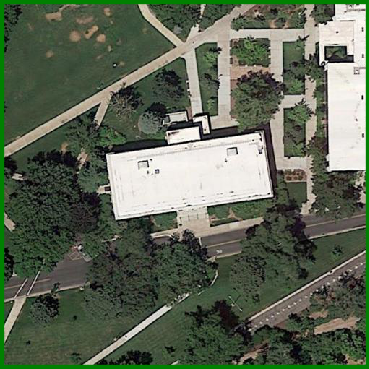} &
		\includegraphics[width=0.09\linewidth]{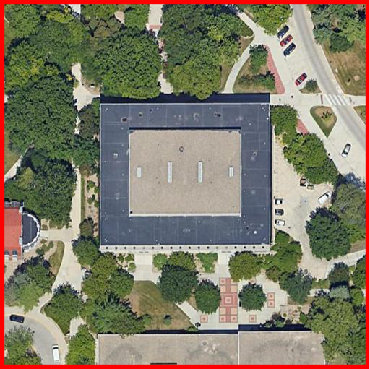} &
		\includegraphics[width=0.09\linewidth]{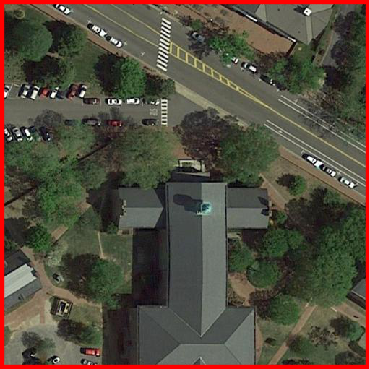} &
		\includegraphics[width=0.09\linewidth]{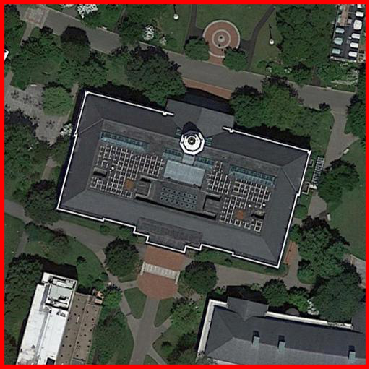} &
		\includegraphics[width=0.09\linewidth]{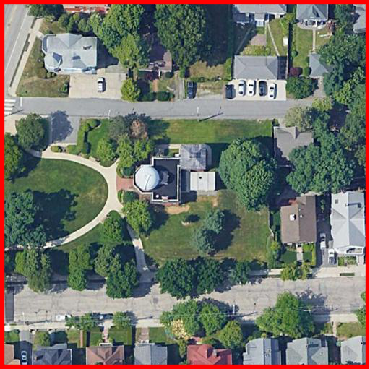} \\
		\multicolumn{11}{c}{(a) Occlusion}\\
		\small{Query} & \multicolumn{5}{l|}{Ground $\rightarrow$ Drone (Top1 $\rightarrow$ Top5)}
		& \multicolumn{5}{l}{Ground $\rightarrow$ Satellite (Top1 $\rightarrow$ Top5)}\\
		\includegraphics[width=0.09\linewidth]{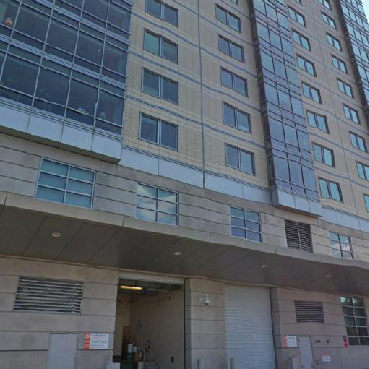} &
		\includegraphics[width=0.09\linewidth]{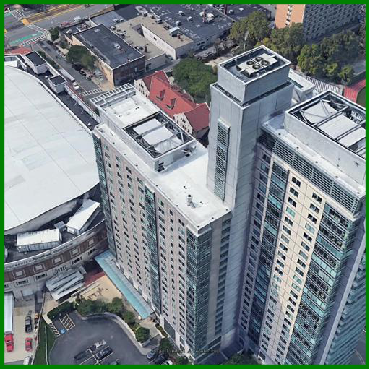} &
		\includegraphics[width=0.09\linewidth]{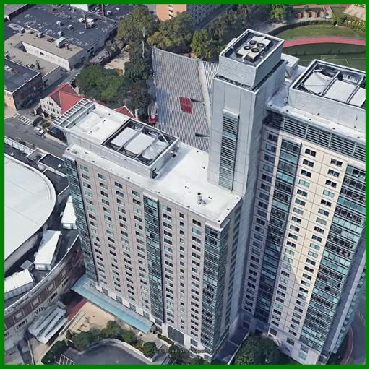} &
		\includegraphics[width=0.09\linewidth]{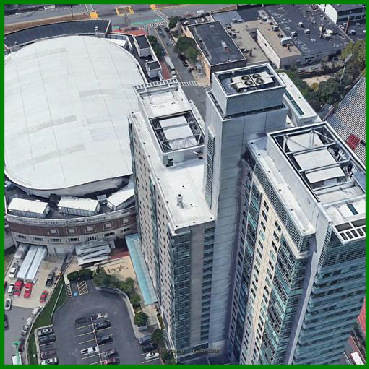} &
		\includegraphics[width=0.09\linewidth]{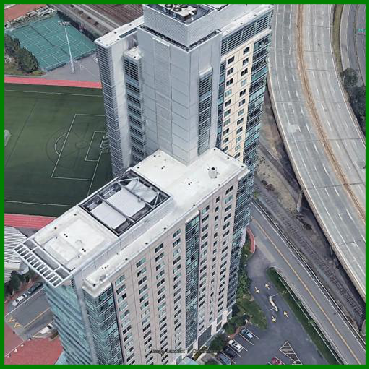} &
		\includegraphics[width=0.09\linewidth]{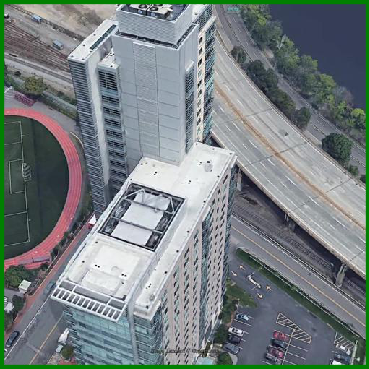} &
		\includegraphics[width=0.09\linewidth]{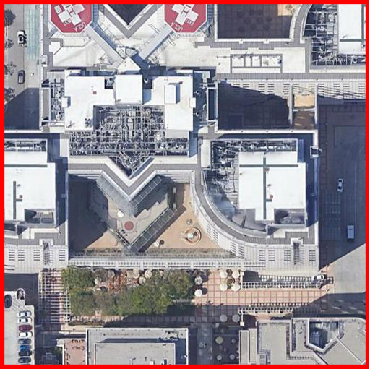} &
		\includegraphics[width=0.09\linewidth]{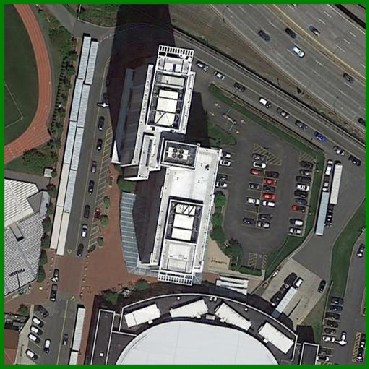} &
		\includegraphics[width=0.09\linewidth]{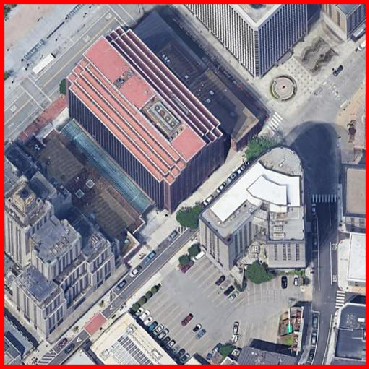} &
		\includegraphics[width=0.09\linewidth]{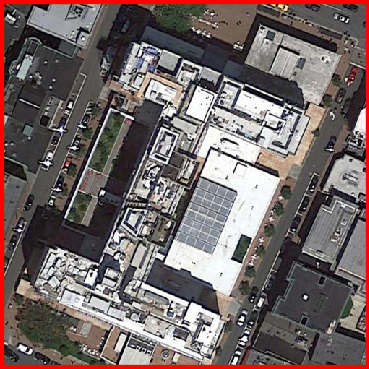} &
		\includegraphics[width=0.09\linewidth]{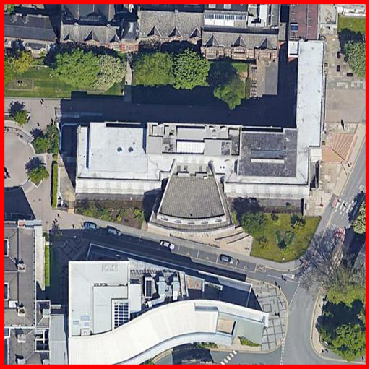} \\
		\includegraphics[width=0.09\linewidth]{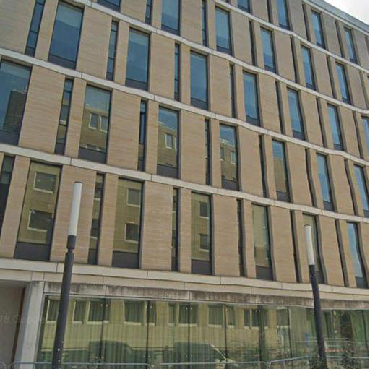} &
		\includegraphics[width=0.09\linewidth]{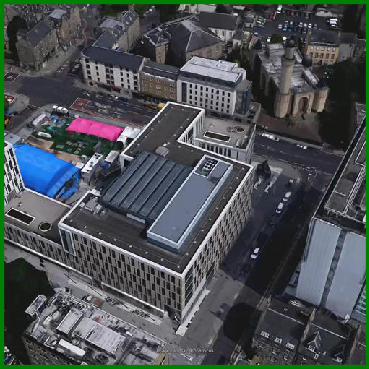} &
		\includegraphics[width=0.09\linewidth]{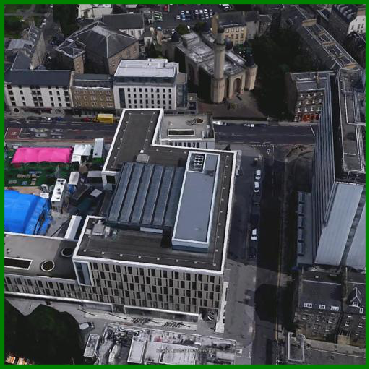} &
		\includegraphics[width=0.09\linewidth]{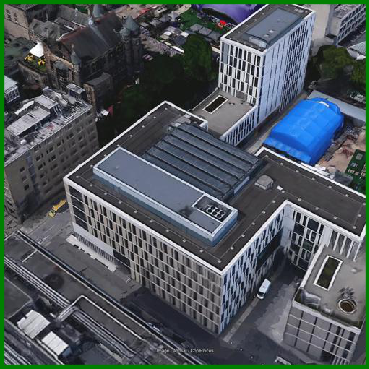} &
		\includegraphics[width=0.09\linewidth]{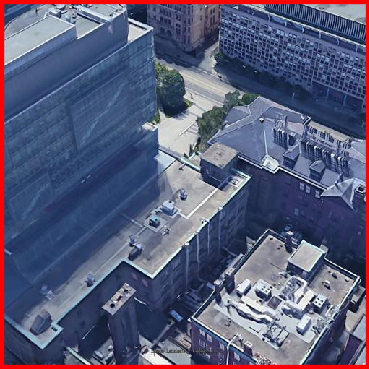} &
		\includegraphics[width=0.09\linewidth]{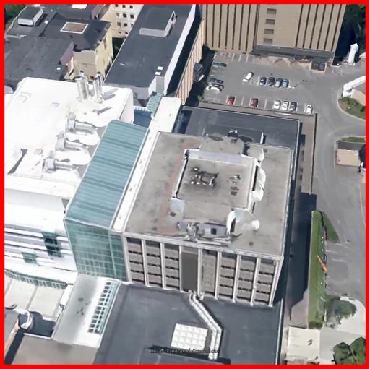} &
		\includegraphics[width=0.09\linewidth]{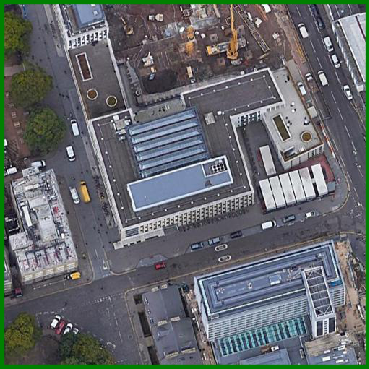} &
		\includegraphics[width=0.09\linewidth]{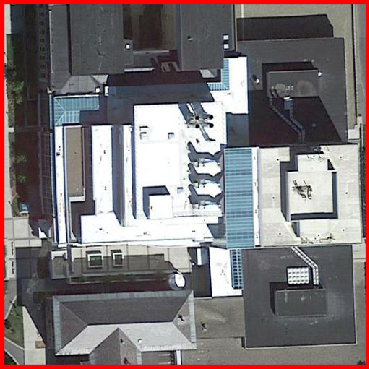} &
		\includegraphics[width=0.09\linewidth]{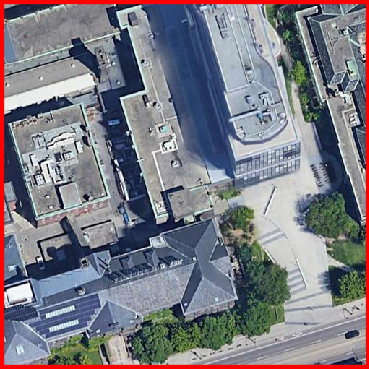} &
		\includegraphics[width=0.09\linewidth]{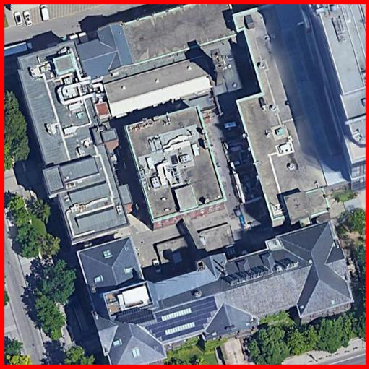} &
		\includegraphics[width=0.09\linewidth]{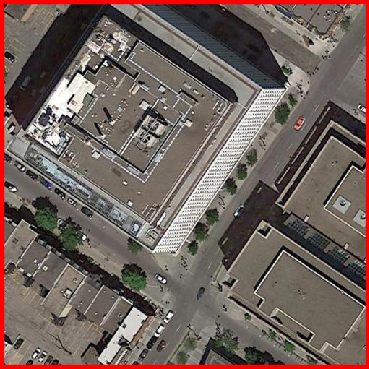} \\\includegraphics[width=0.09\linewidth]{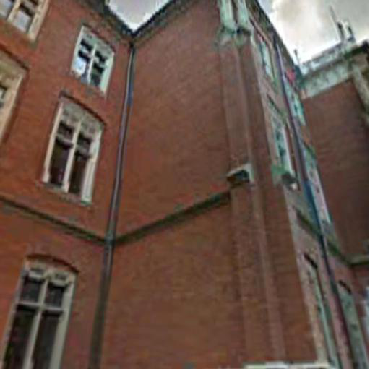} &
		\includegraphics[width=0.09\linewidth]{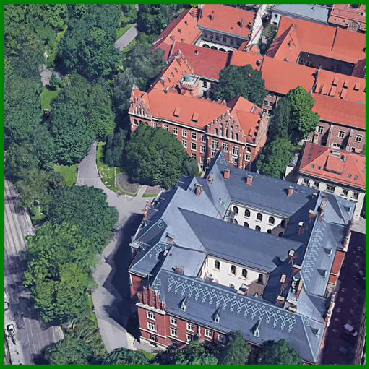} &
		\includegraphics[width=0.09\linewidth]{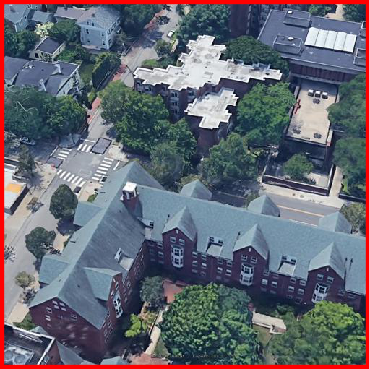} &
		\includegraphics[width=0.09\linewidth]{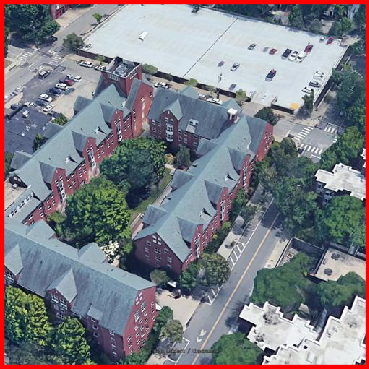} &
		\includegraphics[width=0.09\linewidth]{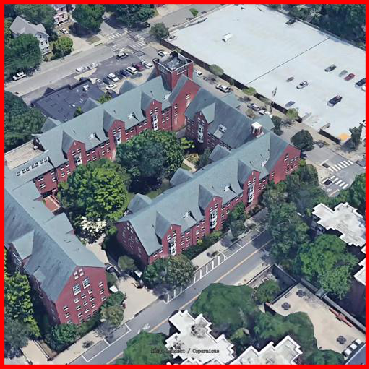} &
		\includegraphics[width=0.09\linewidth]{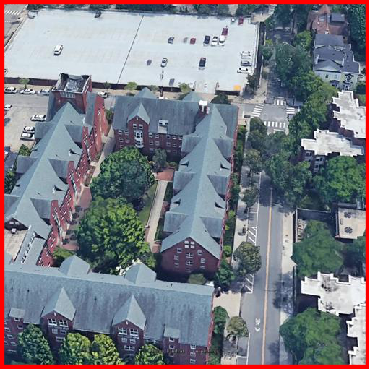} &
		\includegraphics[width=0.09\linewidth]{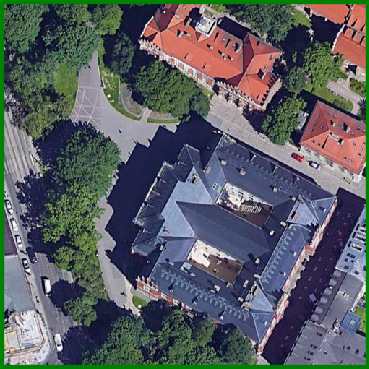} &
		\includegraphics[width=0.09\linewidth]{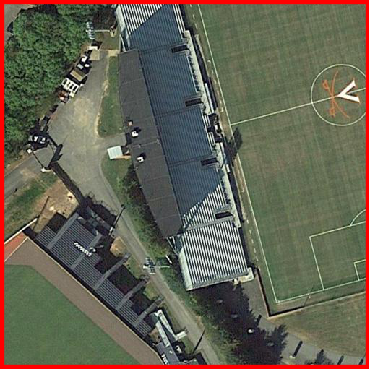} &
		\includegraphics[width=0.09\linewidth]{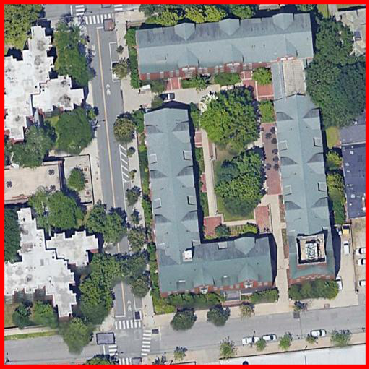} &
		\includegraphics[width=0.09\linewidth]{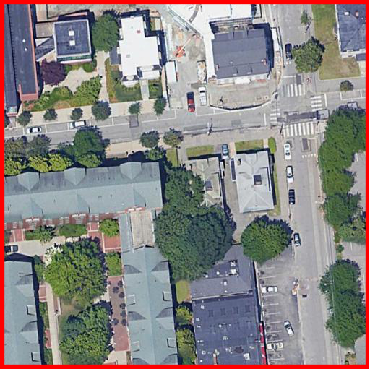} &
		\includegraphics[width=0.09\linewidth]{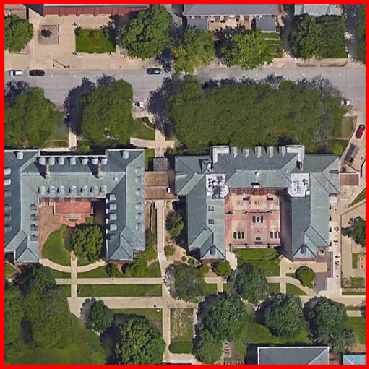} \\
		\multicolumn{11}{c}{(b) Tiny Part}\\
       \end{tabular}
   \caption{The visualization of the ground-to-satellite retrieval results of hard samples by our PLCD framework, including (a) occlusion samples and (b) tiny part samples. From left to right: ground-view query image, the Top 1-5 retrieved drone-view images, and the Top 1-5 retrieved satellite-view images. Green and red borders indicate correct and incorrect retrieved results, respectively.} 
\label{fig:difficult_results}
\end{figure*}

\subsection{Visualization Results and Failure Case}
In this subsection, we visualize more retrieval results. Given a ground-view image, we show its top five drone-view images by Ground$\rightarrow$Drone retrieval, as well as top five satellite-view images by Ground$\rightarrow$Satellite retrieval. The visualization results are shown in Figure~\ref{fig:result}. The top three rows show some good results, we can see that although the viewpoint changes a lot and the landmark is not so obvious in the images, our method can find the satellite images in the top results. 

The bottom two rows also show two failure cases. For the fourth query, we can find some drone-view images, but we fail in finding its satellite-view images. We consider it is because the landmark is a little small and common and does not contain too many special details in the satellite-view images. For the fifth query, the landmark is relatively not so obvious in the ground-view image, making it hard to be found in the drone-view images.

Furthermore, we visualize more retrieval results of the hard samples. Given a hard sample (ground-view image), we show its top five drone-view images by Ground$\rightarrow$Drone retrieval, as well as top-five satellite-view images by Ground$\rightarrow$Satellite retrieval. The visualization results are shown in Figure~\ref{fig:difficult_results}. The top three rows show some results of occlusion, and the bottom three rows show some results of tiny parts. The results demonstrate that our method can be effective in hard samples.

{
\subsection{Computational Efficiency}
For the efficiency, when we train on the University-Earth dataset (2,659 street images, 37,854 drone images and 701 satellite images, single Tesla V100S GPU), the training time of the first step Ground-Drone Cross-view Representation and the second step Satellite-Drone Cross-view Representation are 731$m$$\pm$3$m$38$s$ (Senior Peer: 323$m$18$s$$\pm$1$m$38$s$; Junior Peer: 403$m$$\pm$2$m$) and 87$m$11$s$$\pm$11$s$, respectively. 
}

{
When we test on the University-Earth dataset (2,579 query-street images, 17,119 gallery-drone images and 951 satellite images), the running time of extracting feature and cross-diffusion ranking are 5$m$26$s$$\pm$5$s$ (37$s$ for query images and 4$m$49$s$$\pm$5$s$ for gallery images, single Tesla V100S GPU) and 146$s$$\pm$2$s$ (on CPU), respectively. For each query, the running time of our method is 0.18$s$ in average. All information are shown in Table~\ref{tab:time}. Hence, our method works in real-time application.
\begin{table*}[!h]
    \centering
    \tabcolsep=12pt
    \begin{tabular}{c|c|c}
    \hline
          \multicolumn{2}{c|}{Part} & Time (single Tesla V100S GPU) \\\hline
          \multirow{3}{*}{Ground-Drone Cross-view Representation} & Senior Peer & Training time: 323$m$18$s$$\pm$1m38$s$\\
           & Junior Peer & Training time: 323$m$18$s$$\pm$1$m$38$s$ \\
           & Total & Training time: 403$m$$\pm$2$m$ \\\hline
          \multicolumn{2}{c|}{Satellite-Drone Cross-view Representation} & Training time: 87$m$11$s$$\pm$11$s$ \\ \hline
          \multirow{3}{*}{Cross-View Image Retrieval} & Feature Extraction & Test time: 5$m$26$s$$\pm$5$s$\\
           & Cross-diffusion Ranking & Testing time: 146$s$$\pm$2$s$ (CPU) \\
           & Average & Testing time: 0.18$s$ per query \\
      \hline
    \end{tabular}
    \caption{Results of the training and testing time of PLCD.}
    \label{tab:time}
\end{table*}
}

\section{Conclusion}
In this paper, we argue the defects/challenges of existing approaches in ground-to-satellite geo-localization task, and raise a new cross-view representation and diffusion strategy. {The key novelty of the paper is the idea of exploiting drone-view images in training to improve the performance of ground-satellite image retrieval. As minor contributions, our techniques are also of sufficient novelty. 1) For the framework, we have two independent cross-view networks; 2) For the network architecture, to address the specific mismatch challenge in cross-view image retrieval, by adopting R-MAC and detection layer; 3) For optimization, we adopt peer learning strategy, which is firstly applied to the cross-view image retrieval task; 4) Our diffusion works as a connection between two independent feature spaces, while existing diffusion just runs in single feature space.}

\heading{{Perspective}} {Drone images are not as easy to locate less accurately, compared to satellite images, hence we use satellite images to achieve geo-localization. Meanwhile, the drone-view images for our PLCD do not need annotations or any preparation according to the query. We believe our task fits the realistic application well.} We find that even if we apply a network that achieves a better performance on drone-to-satellite retrieval, it may not necessarily obtain better results on diffusion. What is the relevance between cross-diffusion and cross-view space? We will continuously investigate this point worth nothing. In addition, the external environment also has a significant impact on the ground-to-satellite geo-localization. Small targets~\cite{hu2021capturing} and retrieval in low light~\cite{zeng2020illumination,jiang2020decomposition,jiang2021rain,xu2021exploring} are also worth investigating.

\bibliographystyle{IEEEtran}
\bibliography{egbib}

\begin{IEEEbiography}[{\includegraphics[width=1in,height=1.25in]{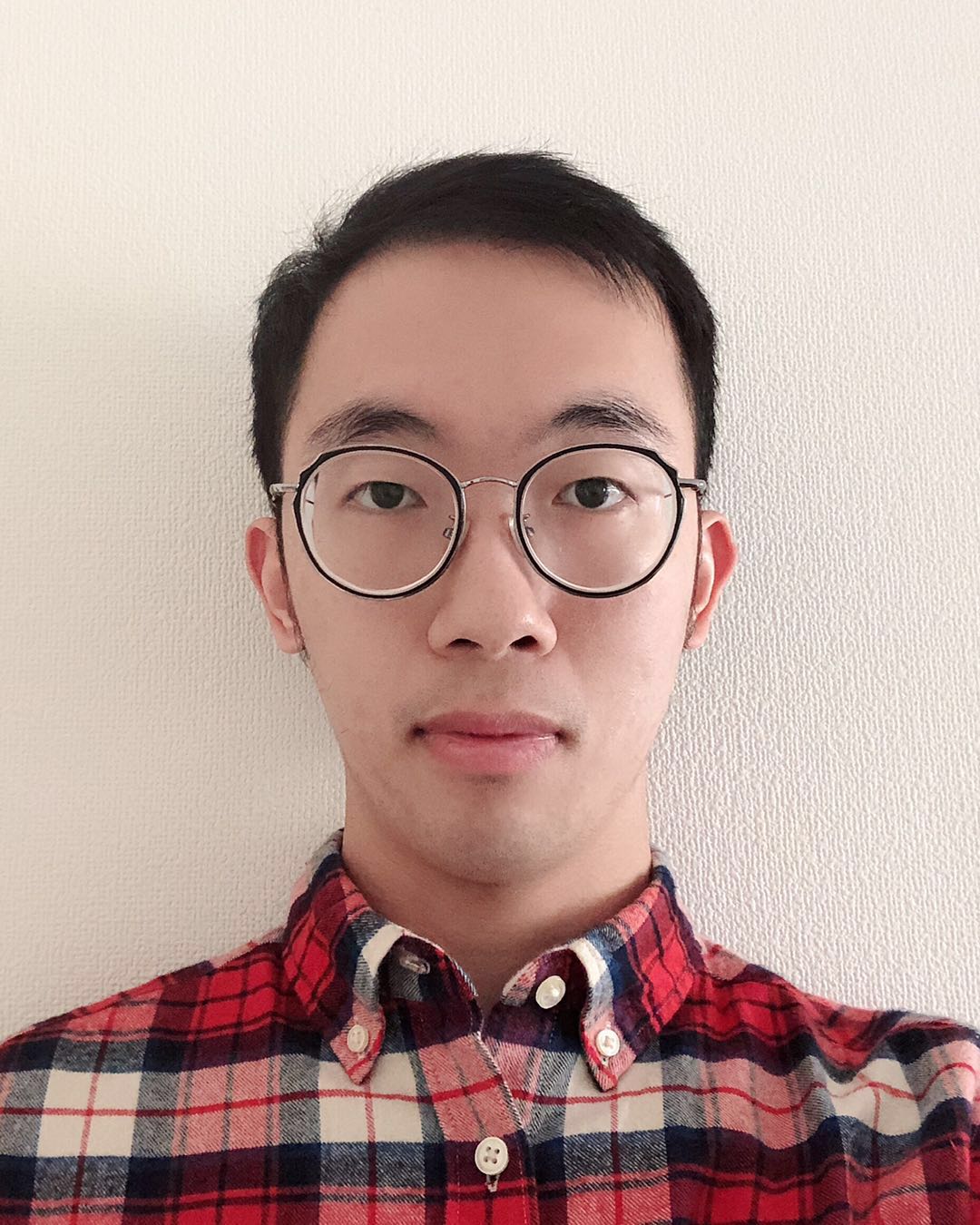}}]{Zelong Zeng} is currently a PhD student at the Department of Information and Communication Engineering, Graduate School of Information Science and Technology, the University of Tokyo, Tokyo, Japan. He receive the M.S degrees from the University of Tokyo in 2020. His research interests include person re-identification and image retrieval. 
\end{IEEEbiography}

\begin{IEEEbiography}[{\includegraphics[width=1in,height=1.25in]{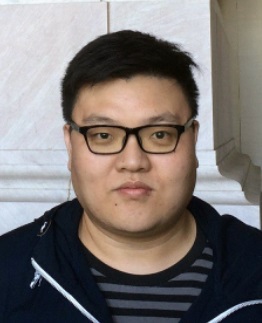}}]{Zheng Wang} (M'19) received the B.S., M.S., and Ph.D. degrees from Wuhan University in 2006, 2008, 2017, respectively. He was a JSPS Fellowship Researcher at the National Institute of Informatics, Japan, and a Project Assistant Professor at The University of Tokyo, Japan. He is currently a Professor at Wuhan University, China. His research interests include person re-identification and instance search. 
\end{IEEEbiography}

\begin{IEEEbiography}[{\includegraphics[width=1in,height=1.25in]{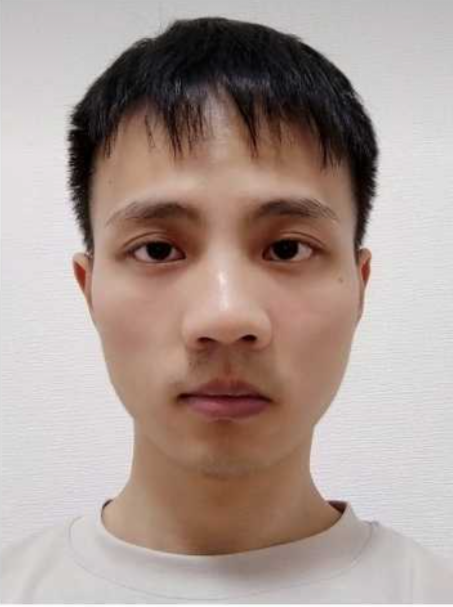}}]{Fan Yang} received the B.E. degree from Zhejiang University in 2015, and the M.E. and Ph.D. degrees from the University of Tokyo in 2018 and 2021, respectively. He is currently a researcher at the National Institute of Informatics, Japan. His research interests include image/video retrieval and person re-identification.
\end{IEEEbiography}

\begin{IEEEbiography}[{\includegraphics[width=1in,height=1.25in]{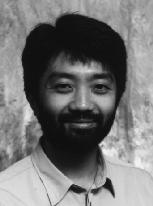}}]{Shin'ichi Satoh} (M'04) received the B.E. degree in electronics engineering and the M.E. and Ph.D. degrees in information engineering from the University of Tokyo, Tokyo, Japan, in 1987, 1989, and 1992, respectively. He has been a Full Professor with the National Institute of Informatics, Tokyo, Japan, since 2004. He was a Visiting Scientist with the Robotics Institute, Carnegie Mellon University, Pittsburgh, PA, USA, from 1995 to 1997. His current research interests include image processing, video content analysis, and multimedia databases.
\end{IEEEbiography}

\end{document}